%% file: memrecon.tex
\documentclass{article}

\PassOptionsToPackage{numbers, compress}{natbib}
\usepackage[main, final]{neurips_2026} % NeurIPS 2026 Main Track
\usepackage{graphicx}
\usepackage{xcolor}
\usepackage{tabularx}
\usepackage{booktabs}
\usepackage{enumitem}
\usepackage{soul}
\definecolor{softamber}{HTML}{FFF1B8}
\sethlcolor{softamber}
\newcommand{\ev}[1]{\hl{#1}}
\usepackage[utf8]{inputenc} % allow utf-8 input
\usepackage[T1]{fontenc}    % use 8-bit T1 fonts
\usepackage{hyperref}       % hyperlinks
\usepackage{url}            % simple URL typesetting
\usepackage{booktabs}       % professional-quality tables
\usepackage{amsfonts}       % blackboard math symbols
\usepackage{nicefrac}       % compact symbols for 1/2, etc.
\usepackage{amsmath}
\usepackage{microtype}      % microtypography
\usepackage{xcolor}         % colors
\usepackage{graphicx}       % figures
\usepackage{array}          % table column helpers
\usepackage{makecell}       % compact multi-line table headers
\usepackage{adjustbox}      % fit wide tables without changing content
\usepackage{threeparttable} % compact table notes
\usepackage{float}
\usepackage[breakable, skins]{tcolorbox} % boxed failure-case examples
\usepackage{fancyvrb}                    % page-breakable verbatim for prompts
\usepackage{fvextra}                    % robust line breaking inside verbatim prompt boxes
\usepackage{pifont}
\usepackage{wrapfig}                    % wraptable for text-wrapped human-validation tables
\usepackage{tikz}
\definecolor{markgreen}{HTML}{00B050}  % vivid green
\definecolor{markyellow}{HTML}{FFC000} % vivid yellow/orange
\definecolor{markred}{HTML}{FF0000}    % vivid red

\DeclareRobustCommand{\fullmark}{%
  \tikz[baseline=-0.55ex]\fill[markgreen] (0,0) circle (0.78ex);%
}

\DeclareRobustCommand{\partmark}{%
  \tikz[baseline=-0.55ex]{%
    \draw[markyellow, line width=0.45pt] (0,0) circle (0.78ex);
    \begin{scope}
      \clip (0,0) circle (0.78ex);
      \fill[markyellow] (-0.78ex,-0.78ex) rectangle (0,0.78ex);
    \end{scope}
  }%
}

\DeclareRobustCommand{\nomark}{%
  \tikz[baseline=-0.55ex, line cap=round]{%
    \draw[markred, line width=1.05pt] (-0.62ex,-0.62ex) -- (0.62ex,0.62ex);
    \draw[markred, line width=1.05pt] (-0.54ex,0.62ex) -- (0.68ex,-0.60ex);
  }%
}

\definecolor{insightblue}{HTML}{0047AB} 

\newtcolorbox{insightbox}{%
  enhanced,
  colback=blue!3,
  colframe=insightblue,
  boxrule=0.8pt,
  arc=1pt,
  left=4pt,
  right=4pt,
  top=1pt,
  bottom=1pt,
  before skip=6pt,
  after skip=4pt,
}

\newtcolorbox{failbox}[1]{%
  enhanced, breakable,
  colback=gray!4, coltitle=black,colbacktitle=gray!15,
  fonttitle=\bfseries\small, title={#1},
  boxrule=0.4pt, arc=2pt, left=4pt, right=4pt, top=2pt, bottom=2pt,
  before skip=4pt, after skip=4pt,
}

\newtcolorbox{promptbox}{%
  enhanced, breakable,
  colback=gray!3, colframe=black!40,
  boxrule=0.35pt, arc=1.5pt,
  left=2pt, right=2pt, top=3pt, bottom=3pt,
  before skip=4pt, after skip=6pt,
}

\newcommand{\bench}{\textsc{MemAudit}}
\newcommand{\code}[1]{\texttt{#1}}
\newcommand{\cat}[1]{\textsc{#1}}

\title{\bench{}: Auditing Long-Term Agent Memory \\ via Hidden User-State Recovery}

\newcommand{\affmark}[1]{\textsuperscript{\normalfont #1}}
\author{%
\begin{tabular}{c}
Enze Ma\affmark{1} \quad
Yufan Zhou\affmark{2} \quad
Wei-Chieh Huang\affmark{1} \quad
Jie Yang\affmark{1} \quad
Huanhuan Ma\affmark{1} \\
Zixuan Wang\affmark{3} \quad
Chengze Li\affmark{1} \quad
Chunyu Miao\affmark{1} \quad
Philip S. Yu\affmark{1} \quad
Zhen Wang\affmark{3} \\
{\normalfont\small
\begin{tabular}{c}
\affmark{1}University of Illinois Chicago \quad
\affmark{2}KU Leuven \quad
\affmark{3}UC San Diego \\
\texttt{ema264@uic.edu, zhw085@ucsd.edu} 
\end{tabular}
}
\end{tabular}
}

\begin{document}

\maketitle

\input{sections/0_abstract}

\input{sections/1_intro}

\input{sections/2_related_work}
\input{sections/3_protocol}
\input{sections/5_results}

\input{sections/6_conclusion}
\input{sections/7_acknowledge}

% \begin{ack}
% 	% Funding and competing-interest disclosures go here for the camera-ready.
% 	% Hidden by the \texttt{ack} environment in the anonymized submission.
% \end{ack}
\medskip

\bibliographystyle{plainnat}
\bibliography{references}

%%%%%%%%%%%%%%%%%%%%%%%%%%%%%%%%%%%%%%%%%%%%%%%%%%%%%%%%%%%%

\clearpage
\appendix
\section*{Appendix}
\addcontentsline{toc}{section}{Appendix}

\input{appendix/limitations}
\input{appendix/broader_impact}
\input{appendix/benchmark_card}
\input{appendix/generation}
\input{appendix/prompts}
\input{appendix/scoring_details}
\input{appendix/human_validation}
\input{appendix/additional_results}
\input{appendix/recall}
\input{appendix/gallery}
\input{appendix/extra_case_studies}

%%%%%%%%%%%%%%%%%%%%%%%%%%%%%%%%%%%%%%%%%%%%%%%%%%%%%%%%%%%%

% \input{checklist.tex}

\end{document}

%% file: sections/0_abstract.tex
\begin{abstract}
Long-term memory promises LLM agents that grow more capable across sessions, maintaining an accurate, evolving understanding of the user that interaction forms. In practice, however, this memory is evaluated mostly through downstream behavior, such as later answers, personalization quality, or task success, which tests that understanding only indirectly and leaves the memory artifact itself largely unaudited. We argue that long-term memory should instead be evaluated as an auditable post-interaction artifact: after ordinary assistance, what structured user state can be reconstructed from the memory the agent leaves behind? We instantiate this view in {\bench{}}, a benchmark in which a memory-equipped agent assists simulated users, each carrying a hidden, taxonomy-anchored user-state bank, across a trajectory of leak-controlled tasks, after which that bank is reconstructed from the agent's resulting memory under both full-store and top-$k$ access. Built on synthetic ground truth for efficient, scalable measurement, \bench{} spans $50$ simulated users with $31$ hidden dimensions each ($1{,}550$ recovery targets) and tests $5$ representative memory systems. Testing state-of-the-art memory agents, we find that successful assistance and recoverable memory behave as distinct capabilities. Task completion nearly saturates, even for a memoryless baseline, while category-balanced recovery stays moderate (about $0.6$) and drops further under top-$k$ retrieval. \bench{} is the first benchmark to study memory recovery directly, reconstructing the user state a system retains and scoring it against ground truth. We see recovery as a concrete objective for future memory agents to optimize, and \bench{} as a step toward an environment where agents are trained to remember their users, growing more faithful the longer they know them~\footnote{Code and data are available at \url{https://github.com/sora1998/MEMAUDIT-bench}.}.
\end{abstract}

%% file: sections/1_intro.tex
\section{Introduction}
\label{sec:intro}

Long-term memory in LLM agents is meant to transform long and fragmented interaction histories into durable user state, the preferences, constraints, skills, knowledge, and prior experiences that should shape how an assistant behaves across sessions. A growing line of systems pursues this goal, from early agents that stored natural-language experiences and reflections \citep{park2023generativeagentsinteractivesimulacra,shinn2023reflexionlanguageagentsverbal,wang2024voyager} to recent designs built around memory banks \citep{zhong2023memorybankenhancinglargelanguage}, evolving notes \citep{xu2025amemagenticmemoryllm}, extracted facts \citep{chhikara2025mem0buildingproductionreadyai}, hierarchical tiers \citep{packer2024memgptllmsoperatingsystems}, and trained memory-operation policies \citep{yue2026memtdensifyingrewardslonghorizon}. Underlying these systems is a shared goal, that with each session the agent should hold a more faithful representation of the user, building toward an assistant that genuinely knows the person it serves.

Progress toward this goal is mostly measured through downstream behavior. Memory-equipped agents are evaluated on the quality of their later answers, their success on follow-up tasks, and the personalization of their responses. The benchmarks behind these measures differ in emphasis. Some probe long-horizon recall and multi-session reasoning, including updating stale facts and forgetting what is outdated \citep{wu2025longmemevalbenchmarkingchatassistants,hu2026evaluatingmemoryllmagents,maharana2024evaluatinglongtermconversationalmemory,he2026memoryarenabenchmarkingagentmemory,chen2026halumemevaluatinghallucinationsmemory,uddin2026recallforgettingbenchmarkinglongterm}, while others probe how closely an assistant's responses match a user's profile, preferences, and history \citep{salemi2024lamplargelanguagemodels,kumar2024longlampbenchmarkpersonalizedlongform,jiang2025knowmerespondme,jiang2025personamemv2personalizedintelligencelearning,zhao2025personalensbenchmarkpersonalizationevaluation,xiao2026alpsbenchllmpersonalizationbenchmark,jiayang2026amemgyminteractivememorybenchmarking,liu2026permabenchmarkingpersonalizedmemory}. Across this variety, the score still rests on the agent's downstream behavior, predominantly its task success and output quality (Figure~\ref{fig:concept}, left), as Table~\ref{tab:benchmark_comparison} details. On these measures, recent memory systems report strong results \citep{chhikara2025mem0buildingproductionreadyai,xu2025amemagenticmemoryllm,yue2026memtdensifyingrewardslonghorizon}. However, a high task-success score is a weak diagnostic of memory quality. A correct response can come from a retrieved span, the surrounding context, or a lucky guess, none of which requires a durable model of the user; in our experiments, a memoryless agent already attains near-perfect task success. Task success and memory formation are therefore distinct properties, and strong performance on the former need not imply that any user model was formed. Assessing memory thus calls for a different yardstick. We propose \emph{recovery} (Figure~\ref{fig:concept}, right), treating memory as an auditable post-interaction artifact and asking how much of a user's hidden state can be reconstructed.

\input{figures/fig_concept}
We realize this idea in \textbf{\bench{}}, which probes long-term agent memory through hidden user-state recovery. Each user in \bench{} is a simulator built around a hidden, taxonomy-anchored bank of user-state targets, spanning skills, knowledge, episodic events, a self-model, and assistance preferences, such as a standing preference for blunt feedback or an early experience that still shapes how the user decides. This bank stays hidden from the agent and serves only as ground truth. Each target is paired with a leak-controlled task that draws out the relevant evidence during ordinary help while keeping the target unnamed, so a system can score well only by forming the memory itself. After the agent works through a user's tasks and writes to its store, we probe what the memory retains, reconstructing each hidden target under two access modes: a full-store \emph{dump\_all} probe and the system's own top-$k$ \emph{retrieve} interface. These two modes separate whether evidence was ever written from whether it can still be reached. We release the full benchmark, $50$ simulated users with $31$ hidden user-state dimensions each ($1{,}550$ recovery targets), together with all prompts, evaluation traces, and pipeline code. Figure~\ref{fig:concept} summarizes this framing.

We benchmark five representative memory systems with \bench{}, auditing what each preserves about $50$ simulated users across $1{,}550$ hidden user-state targets. Our experiments mainly surface three findings, each marking where long-term memory can advance next:

\begin{itemize}[leftmargin=*, itemsep=0pt, topsep=0pt]
  \item \textbf{Task success is not enough to measure memory.} Agents can complete nearly every task, even without memory, so behavioral scores can no longer separate good memory from none. Recovery reopens a measurable axis of progress, namely how much of the user a system preserves, the quantity long-term memory was meant to improve.
  \item \textbf{A major bottleneck is retrieval-aware consolidation.} Systems capture the raw interaction yet seldom distill it into compact user state, so what they hold is hard to surface later. The lever for better memory is a write policy that turns fleeting evidence into durable, reusable claims, which capacity and retrieval tuning alone leave untouched.
  \item \textbf{The open frontier is episodic, relational memory.} Stable preferences are largely handled, while one-off experiences stay hard, because recovering them means binding an event to the context and consequence that gave it meaning. This binding is what lets an assistant genuinely know a person, and aggregate recovery hides it, so future designs should target episodic structure directly.
\end{itemize}

Taken together, these findings recast long-term memory as something to measure and build directly at the level of the artifact. By making that artifact auditable, \bench{} gives the field a direct handle on it, and we hope it moves us toward agents that accumulate a faithful model of the user and grow more useful the longer they know someone.

\input{tables/tab_benchmark_comparison}

%% file: figures/fig_concept.tex
\begin{wrapfigure}{r}{0.55\textwidth}
    \centering
    \includegraphics[
    width=0.55\textwidth,
    trim=2 138 69 84,
    clip
    ]{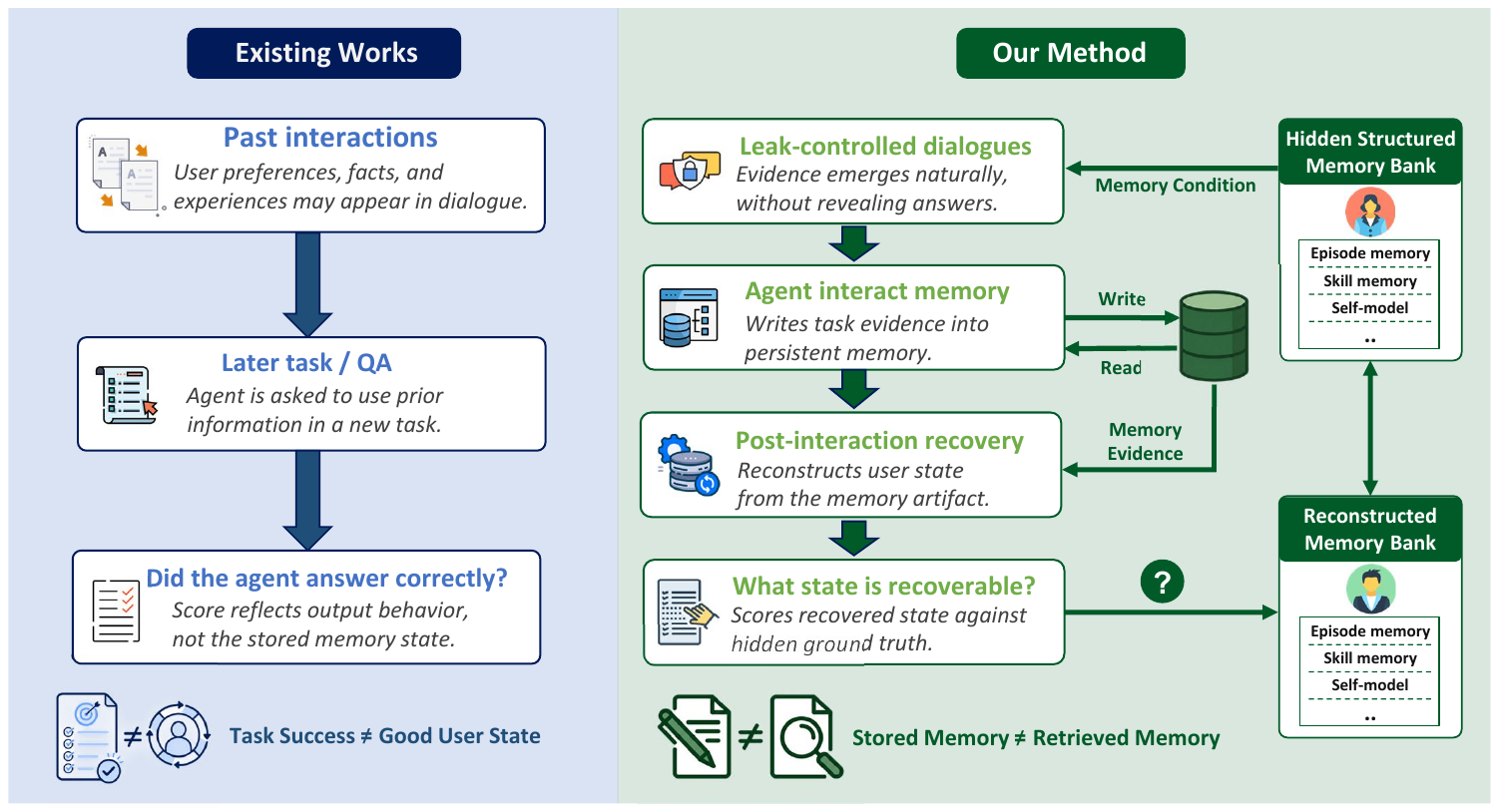}
    \caption{
    Behavior-based memory evaluation (left) versus the \bench{} recovery audit (right).
    }
    \label{fig:concept}
\end{wrapfigure}

%% file: tables/tab_benchmark_comparison.tex
\begin{table}[htbp]
\caption{
Comparison with representative long-term memory and personalization benchmarks.
\textit{User-Level Trajectory}: repeated interactions belong to the same user.
\textit{Dynamic Interaction}: evaluation-time interaction yields new evidence.
\textit{Hidden User Bank}: pre-defined user-state ground truth is hidden.
\textit{Recovery-Based Scoring}: recovered state is scored against ground truth.
\textit{Memory as Source}: scores are based on the system's memory.
\textit{Write--Read Split}: written storage and retrieval are separated.
Markers indicate full, partial, or absent support.
}
\label{tab:benchmark_comparison}
\centering
\small
\setlength{\tabcolsep}{5pt}
\renewcommand{\arraystretch}{1.0}
\begin{adjustbox}{max width=\linewidth}
\begin{tabular}{@{}lcccccc@{}}
\toprule
Method &
\makecell{User-Level\\Trajectory} &
\makecell{Dynamic\\Interaction} &
\makecell{Hidden\\User Bank} &
\makecell{Recovery-Based\\Scoring} &
\makecell{Memory\\as Source} &
\makecell{Write--Read\\Split} \\
\midrule

LongMemEval~\citep{wu2025longmemevalbenchmarkingchatassistants}
& \fullmark & \nomark & \nomark & \nomark & \nomark & \nomark \\

MemoryAgentBench~\citep{hu2026evaluatingmemoryllmagents}
& \fullmark & \fullmark & \nomark & \nomark & \nomark & \nomark \\

MemoryArena~\citep{he2026memoryarenabenchmarkingagentmemory}
& \fullmark & \fullmark & \nomark & \nomark & \nomark & \nomark \\

LaMP / LongLaMP~\citep{salemi2024lamplargelanguagemodels,kumar2024longlampbenchmarkpersonalizedlongform}
& \partmark & \nomark & \nomark & \nomark & \nomark & \nomark \\

AlpsBench~\citep{xiao2026alpsbenchllmpersonalizationbenchmark}
& \fullmark & \nomark & \nomark & \partmark & \nomark & \nomark \\

PersonaMem-v2~\citep{jiang2025personamemv2personalizedintelligencelearning}
& \fullmark & \nomark & \partmark & \nomark & \nomark & \nomark \\

AMemGym \citep{jiayang2026amemgyminteractivememorybenchmarking}
& \fullmark & \fullmark & \fullmark & \nomark & \nomark & \partmark \\

\midrule
\textbf{\bench{}}

& \fullmark & \fullmark & \fullmark & \fullmark & \fullmark & \fullmark \\

\bottomrule
\end{tabular}
\end{adjustbox}
\end{table}

%% file: sections/2_related_work.tex
\section{Related Work}
\label{sec:related}

\noindent \textbf{Memory systems for LLM agents.}
Agent-memory systems differ in how they turn interaction history into persistent state. Early systems stored natural-language experiences and reflections to support believable behavior and self-improvement \citep{park2023generativeagentsinteractivesimulacra,shinn2023reflexionlanguageagentsverbal}, while other agents maintained reusable skill libraries for open-ended learning \citep{wang2024voyager}. More recent systems make long-term memory explicit through context and archival tiers \citep{packer2024memgptllmsoperatingsystems,hu2026pancake}, persistent user memories \citep{zhong2023memorybankenhancinglargelanguage}, extracted and consolidated conversational memories \citep{chhikara2025mem0buildingproductionreadyai}, evolving linked notes \citep{xu2025amemagenticmemoryllm}, and learned memory-operation policies \citep{yue2026memtdensifyingrewardslonghorizon}. These systems differ in what they retain and in how they write, consolidate, organize, and retrieve it. Their progress is typically read off downstream behavior, which leaves unmeasured how much of the user each design actually preserves. \bench{} measures exactly that, auditing the recoverable user state in the artifact each mechanism leaves behind.

\noindent \textbf{Long-horizon, retrieval, and retention benchmarks.}
Long-memory benchmarks commonly evaluate whether models can recall or use information from previous interactions. Very long-term conversational-memory evaluation tests recall over extended multi-session interactions \citep{maharana2024evaluatinglongtermconversationalmemory}, while LongMemEval evaluates chat assistants on event summarization, multi-session reasoning, temporal reasoning, knowledge updates, and abstention \citep{wu2025longmemevalbenchmarkingchatassistants}. MemoryAgentBench broadens this view to retrieval, test-time learning, long-range understanding, and selective forgetting \citep{hu2026evaluatingmemoryllmagents}; MemoryArena further couples memory with action in interdependent multi-session agentic tasks \citep{he2026memoryarenabenchmarkingagentmemory}. Other work tests hallucination during memory extraction, update, and question answering \citep{chen2026halumemevaluatinghallucinationsmemory}, or penalizes obsolete-memory use under evolving user histories \citep{uddin2026recallforgettingbenchmarkinglongterm}. These benchmarks sharpen how we evaluate memory operations and memory-conditioned behavior, yet their scores stay tied to answers, actions, retrieval hits, or operation correctness, any of which a system can get right without forming a durable user model. \bench{} reads the artifact directly, reconstructing hidden user state under both full-store and retrieval access to separate what was written from what is reachable.

\noindent \textbf{Personalization and persona-conditioned evaluation.}
Personalization benchmarks evaluate whether assistants adapt outputs to a user's preferences, traits, or history, extending a line of task-oriented and user-centered dialogue systems \citep{mo2023rollupsleeves,chen2021bootstrapping}. LaMP \citep{salemi2024lamplargelanguagemodels} evaluates personalized language modeling from user profiles and histories, and LongLaMP \citep{kumar2024longlampbenchmarkpersonalizedlongform} extends this setting to personalized long-form generation. KnowMe \citep{jiang2025knowmerespondme} studies dynamic user profiling and personalized response generation, while PersonaMem-v2 \citep{jiang2025personamemv2personalizedintelligencelearning} considers implicit personas, long-context interaction histories, and agentic memory for compact personalization. AMemGym \citep{jiayang2026amemgyminteractivememorybenchmarking} evaluates on-policy assistants with hidden evolving user states and write/read-style diagnostics, but still using downstream personalized QA and state-query outputs as the main evaluation signals.
PersonaLens\citep{zhao2025personalensbenchmarkpersonalizationevaluation} evaluates personalization through structured user-memory lenses; AlpsBench \citep{xiao2026alpsbenchllmpersonalizationbenchmark} studies real-dialogue memorization and preference alignment; and PERMA \citep{liu2026permabenchmarkingpersonalizedmemory} evaluates personalized memory agents in event-driven preference and task environments. These benchmarks move closer to long-term assistant settings, yet they grade memory through downstream outputs, whether personalized answers, preference alignment, or state queries. \bench{} keeps the user state hidden during the interaction and scores whether the memory left behind can reconstruct it, making the user model itself the measured quantity.

%% file: sections/3_protocol.tex
\section{The \bench{} Protocol}
\label{sec:protocol}
\input{figures/fig_protocol}

Figure~\ref{fig:protocol} summarizes the \bench{} pipeline. We formally define the memory recovery problem (Section~\ref{sec:formulation}), construct simulated users with hidden ground truth and leak-controlled tasks (Section~\ref{sec:data_construction}), run them against a memory-equipped agent (Section~\ref{sec:episode}), reconstruct and score the hidden user state from the memory left behind (Section~\ref{sec:scoring}), and attribute low-recovery cases to their cause (Section~\ref{sec:attribution}).

\subsection{Problem Formulation}
\label{sec:formulation}

We write a user's hidden state as $u = (u_1, \dots, u_{31}) \in \mathcal{U}$, a vector of $31$ target dimensions drawn from a fixed taxonomy (Section~\ref{sec:data_construction}), where $\mathcal{U}$ is the space of such states. The agent never sees $u$, which exists only as ground truth for scoring. Auditing memory amounts to asking how much of $u$ survives three lossy steps, each introducing its own notation. \emph{Expose} turns the hidden state into evidence $e$, the part of $u$ the user reveals while being helped. \emph{Collect} turns that evidence into a stored artifact $m_{\text{final}}$, the part the system chooses to write and keep. \emph{Recover} turns the stored artifact back into an estimate $\hat{u}$ of the hidden state. Compactly,
\[
  u \;\xrightarrow{\;\text{expose}\;}\; e \;\xrightarrow{\;\text{collect}\;}\; m_{\text{final}} \;\xrightarrow{\;\text{recover}\;}\; \hat{u}.
\]
Every arrow can lose information, and \bench{} is built to localize where the loss occurs. These three steps also organize the rest of the pipeline. The simulation rollout (Section~\ref{sec:episode}) carries out \emph{expose} and \emph{collect}, recovery scoring (Section~\ref{sec:scoring}) carries out \emph{recover}, and failure attribution (Section~\ref{sec:attribution}) localizes which arrow lost a given target. The three steps below make each arrow precise.

\textbf{Expose.} Over a trajectory of $T$ tasks, a user simulator $\pi_u$, a policy conditioned on the hidden state $u$, interacts with a memory-equipped agent. The dialogue never states $u$ outright; instead, at each turn $t$ the simulator samples evidence $e_t \sim \pi_u(\,\cdot \mid u, c_t)$ that partially reveals it, where $c_t$ is the local context of the current task and dialogue history. The evidence $e$ in the diagram above is what these turns disclose across the trajectory. Whether a given dimension $u_i$ is ever exposed depends on the leak-controlled task and the interaction.

\textbf{Collect.} As the evidence arrives, the agent's memory policy $\pi_{\text{mem}}$ chooses an operation $o_t \in \mathcal{O}$, the set of memory operations the system supports, such as writing, merging, or revising an entry. The operation is sampled $o_t \sim \pi_{\text{mem}}(\,\cdot \mid m_{t-1}, e_t)$ from the current store $m_{t-1}$ and the new evidence, and an update rule $\mathcal{T}$ applies it to form the next store $m_t = \mathcal{T}(m_{t-1},\, e_t,\, o_t)$, starting from an empty store $m_0$. Systems differ in how they instantiate $\pi_{\text{mem}}$ and $\mathcal{O}$ (Section~\ref{sec:setup}). After all $T$ tasks, the final store $m_{\text{final}} = m_T$ reflects the memory design and varies across systems even on identical evidence.

\textbf{Recover.} Recovery is measured per dimension. For each dimension $i$, a read operator $\mathcal{R}$ first surfaces memory for a fixed per-dimension query $q_i$, and a slot-filling reader $\rho$ turns the returned memory into an estimate $\hat{u}_i = \rho\bigl(\mathcal{R}(m_{\text{final}}, q_i)\bigr)$ of the target; a judge then scores $\hat{u}_i$ against $u_i$, and the recovery score $B$ aggregates these per-dimension judgments (Section~\ref{sec:scoring}). \bench{} probes two read operators. The full-store operator $\mathcal{R}_{\text{dump}}(m, q) = m$ ignores the query and returns the entire store, and the retrieval operator $\mathcal{R}_{\text{retr}}(m, q)$ returns the top-$k$ items the system's search interface retrieves for $q$; these are the \code{dump\_all} and \code{retrieve} access modes. We use two operators because a memory system can fail in two distinct places: it can fail to write the evidence at all, or it can write the evidence yet fail to surface it on demand. Reading the full store probes the first, reading through the system's own search probes the second, so comparing the two tells us whether a low recovery is a write-side or a read-side failure.

\subsection{Simulated Users and Leak-Controlled Tasks}
\label{sec:data_construction}

Auditing recovery needs something ordinary evaluations lack: a known, structured ground truth for each user that the agent cannot simply read off the dialogue. We therefore evaluate against simulated users. The choice is deliberate, because only a synthetic user lets us fix the hidden state in advance, control exactly what is and is not disclosed during help, and scale to many users and dimensions without privacy or annotation cost. User simulation itself is a standard tool \citep{wang2026mind2dialoguetraininghumanawarelanguage}; what matters here is what our simulator carries. Each simulated user is defined by two hidden artifacts, generated once and then held fixed:

\begin{itemize}[leftmargin=*, itemsep=0pt, topsep=0pt]
  \item \textbf{A persona profile} that gives the user a coherent identity to act from. We generate it with DeepPersona \citep{wang2025deeppersonagenerativeenginescaling}, a generative engine that produces rich, internally consistent synthetic personas at scale, with detailed background, traits, and circumstances. This depth matters for our setting, because a detailed profile lets the simulator hold stable preferences, knowledge, and voice across dozens of otherwise unrelated tasks, so the evidence stays faithful to who they are over a long trajectory.
  \item \textbf{A taxonomy-anchored memory bank} of $31$ target dimensions in five categories (\cat{Skill Memory}, \cat{Knowledge Memory}, \cat{Episodic Memory}, \cat{Self Model}, \cat{Assistance Preference}). Each dimension is seeded from an established source, so the targets rest on prior constructs: O*NET work and knowledge descriptors \citep{Online2026-ci}, Conway's autobiographical-memory taxonomy and McAdams' life-story constructs \citep{CONWAY2005594,McAdams1997TheSW}, and HEXACO facets \citep{Ashton2007-vl,Lee2018-jf}. This bank is the hidden ground truth and carries no psychometric claim.
\end{itemize}

With the hidden state fixed, the second half of construction generates, for each dimension, the assistance task that gives it a chance to surface. The central difficulty is leakage. A task that asks about the target directly would make recovery trivial and uninformative, so the task must invite the evidence without ever naming it. Two mechanisms enforce this:
\begin{itemize}[leftmargin=*, itemsep=0pt, topsep=0pt]
  \item \textbf{Indirect task design.} Each task requests ordinary help on a topic adjacent to the target dimension and never states the target value.
  \item \textbf{A blind anti-fishing critic} that sees only the task statement and rejects or rewrites any task whose answer is already recoverable from that statement alone.
\end{itemize}
The result is a task set that creates natural openings for disclosure while keeping the hidden bank out of the initial request. Full bank schema, taxonomy mapping, and the generation and anti-fishing prompts are given in Appendices~\ref{app:generation} and~\ref{app:prompts}.

\subsection{Simulation Rollout}
\label{sec:episode}

Section~\ref{sec:data_construction} builds the simulator and its tasks; this step runs them. Each accepted task becomes a multi-turn episode in which the simulated user and a memory-equipped agent work together, and the agent reads from and writes to its memory as the conversation proceeds. This is where evidence is produced and where memory is formed, so the rollout is built around one goal: to hold everything except the memory module constant so that any difference in the resulting memory traces is due to memory design alone.

\textbf{Experimental controls.} The task set, simulator prompt, dialogue protocol, and assistant scaffold are identical across systems, and only the memory module changes. Memory persists across all $T$ of a user's tasks and resets before the next user, so each user is a clean trajectory. The agent sees only the current task, the dialogue so far, and whatever its memory returns, and never the hidden bank. An episode ends when the simulator marks the task \code{satisfied} or a turn cap is reached. One subtlety remains. Because each system's memory reads shape its later replies, transcripts can diverge across systems even under identical task conditions; we treat this divergence as part of the memory-conditioned loop under test, and rely on failure attribution (Section~\ref{sec:attribution}) to separate genuine memory failures from cases where the evidence never surfaced. Simulator and episode-control prompts are in Appendix~\ref{app:prompts}.

\subsection{Recovery Scoring}
\label{sec:scoring}

Once a user's trajectory is complete, scoring asks how much of the hidden bank can be reconstructed from the memory the system left behind:
\begin{itemize}[leftmargin=*, itemsep=0pt, topsep=0pt]
  \item \textbf{Grading a reconstruction.} A dimension's target is a short natural-language claim, for which exact string match is too brittle and free generation too permissive. We reconstruct each dimension with a slot-filling reader and grade it with a separate LLM judge on a five-level scale $\{0, 0.25, 0.5, 0.75, 1.0\}$ against the hidden target (rubric in Appendix~\ref{app:scoring_details}).
  \item \textbf{Aggregating across categories.} The five categories hold different numbers of dimensions, so a plain mean would let the largest dominate. We average within each category and then average the five category means with equal weight, giving the category-balanced recovery score $B$.
  \item \textbf{Scoring under two access modes.} We compute $B$ under both read operators from Section~\ref{sec:formulation}. \code{dump\_all} exposes the full written store, so $B$ measures whether the evidence was preserved anywhere; \code{retrieve} exposes only the top $k{=}5$ items the system returns for a fixed, per-dimension query, so $B$ measures whether that evidence is reachable through the normal interface. \code{dump\_all} is deliberately not budget-aware; when a store overflows the reader's context, the resulting drop is itself a diagnostic about auditability, reported separately from the comparable scores.
\end{itemize}

Beyond recovery, we report task fit $A$, mean turns $C$, preference score $D$, and memory footprint $E$. Appendices~\ref{app:scoring_details} and~\ref{app:prompts} detail every metric, the serialization formats, and the prompts given~to~each~judge.

\subsection{Failure Attribution}
\label{sec:attribution}

When recovery is low, we use failure attribution to find out exactly what went wrong. A low score is not always the memory system's doing, since the target may never have been invited by the task, raised by the assistant, or disclosed by the simulator. Attribution assigns each low-scoring case to its true cause, so that only genuine memory faults are counted as such. Cases scoring at least $0.75$ are labeled \code{ok} and skip attribution; the rest pass through three stages. Full attribution prompts, decision rules, and audited examples appear in Appendices~\ref{app:prompts}, \ref{app:scoring_details}, and~\ref{app:gallery}.

\begin{itemize}[leftmargin=*, itemsep=0pt, topsep=0pt]
  \item \textbf{Task-design check (Stage~0).} A task-design oracle checks, from the task statement and ground truth alone, whether the task could plausibly invite the target. If it could not, the low score reflects the task, and we record it as a task-design failure (\code{task\_design\_failure}). Oracle verdicts are cached per $(\text{user}, \text{dimension})$ and shared across runs.
  \item \textbf{Disclosure check (Stage~1).} Otherwise, a judge reads the full transcript. If the dialogue carried enough evidence to reconstruct the target, the memory had its chance and missed it, a genuine memory failure (\code{memory\_failure}).
  \item \textbf{Elicitation versus strictness (Stage~2).} If the evidence never surfaced, the case is split into \code{agent\_elicitation\_failure} (the assistant never approached the topic) and \code{simulator\_too\_strict} (the topic arose, but the simulator did not elaborate).
\end{itemize}

%% file: figures/fig_protocol.tex
\begin{figure}[hbp]
    \centering
    \includegraphics[
    width=0.85\linewidth,
    trim=0 200 240 85,
    clip
    ]{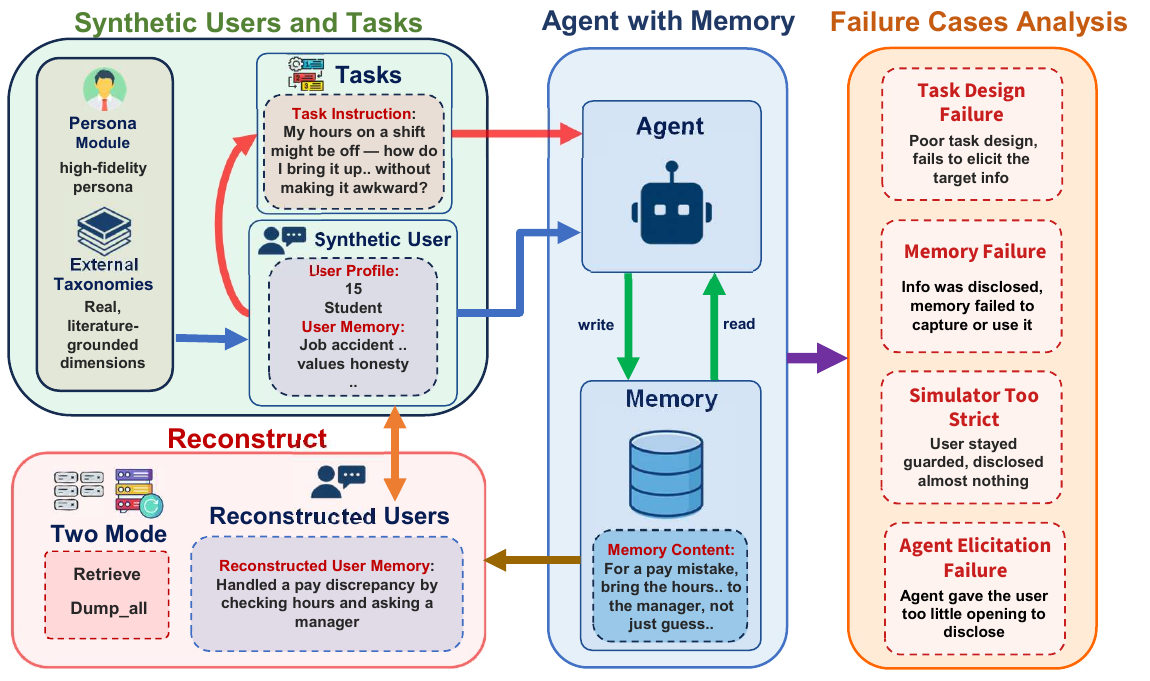}
    \caption{
        The \bench{} pipeline: interaction, \emph{dump\_all}/\emph{retrieve} recovery, and failure attribution.
        }
    \label{fig:protocol}
\end{figure}

%% file: sections/5_results.tex
\section{Experiments and Results}
\label{sec:results}

The results show that long-term memory failures are not visible from immediate task success alone. Across systems, agents can complete the local task and satisfy the simulator while still failing to leave behind a recoverable user-state artifact. We therefore read memory as a lifecycle, in which evidence must be written, turned into user state, surfaced through the read interface, and preserved across kinds of user state. Tables~\ref{tab:main} and~\ref{tab:main_percat} give the main quantitative evidence, while Table~\ref{tab:attrib} and the audited cases diagnose where missing reconstructions arise.

\subsection{Experimental Setup}
\label{sec:setup}

\noindent \textbf{Memory systems compared.}
We compare five memory settings under the \bench{} protocol. \code{nomem} disables memory and provides a no-memory baseline. \code{longctx\_full} stores every turn verbatim and uses sentence-transformer cosine retrieval in retrieve mode, providing a raw evidence-preservation baseline. \code{amem} follows A-Mem \citep{xu2025amemagenticmemoryllm}, storing LLM-written note units with linking, Zettelkasten-style organization, and memory evolution. \code{mem0} follows Mem0 \citep{chhikara2025mem0buildingproductionreadyai}, extracting and consolidating conversational memories with retrieval over the resulting store. \code{memt} follows Mem-T \citep{yue2026memtdensifyingrewardslonghorizon}, using Mem-T-4B as a trained memory-operation policy over a lightweight hierarchical memory database. Together, these settings span no memory, raw-turn storage, evolving note memory, extracted conversational memory, and a trained memory-operation policy.

\noindent \textbf{Configuration.}
The benchmark run uses the \(N{=}50\), \(T{=}31\), 25-turn configuration. The slot-filling evaluator, recovery judge, disclosure judge, and attribution oracle use \code{gpt-5.4-mini}; \code{memt} is served locally via vLLM. The release includes fixed benchmark instances and full evaluation traces: banks, task prompts, dialogue histories, memory stores, slot-fill predictions, recovery and preference judge outputs, attribution labels, and rationales. These artifacts make aggregate metrics recomputable and allow individual scores to be traced to the underlying task, transcript, and memory evidence; prompt templates, scoring rubrics, and attribution decision rules appear in Appendices~\ref{app:prompts} and~\ref{app:scoring_details}. To test whether the main findings depend on evaluator choice or scoring-prompt wording, we also perform frozen rescoring with alternative recovery judges and rubric-preserving prompt paraphrases; the headline orderings are preserved, with changes confined to original near-ties below $0.03$ (Appendix~\ref{app:robustness_checks}).

\input{tables/tab_main}

\subsection{Does Task Success Imply a Good Memory?}
In Table~\ref{tab:main}, all systems nearly saturate the immediate assistant objective, with task completion at $99.87\%$--$99.94\%$, preference scores of $4.58$--$4.66$/5, and only $2.4$--$2.5$ turns per episode on average. Even \code{nomem} achieves near-perfect task completion, showing that local task success is not the discriminating signal. The separation appears only when we ask whether the hidden user bank can be reconstructed from what the system kept. Under \code{dump\_all}, \code{amem}, \code{longctx\_full}, and \code{mem0} reach only moderate recovery scores ($B{=}0.611$--$0.624$), and under \code{retrieve} they fall further to $B{=}0.473$--$0.540$. Successful interaction can thus leave behind a memory artifact that is incomplete or out of operational reach.
\begin{insightbox}
\textbf{Finding 1.} Completing tasks well does not mean a system has preserved recoverable user state. Good in-the-moment assistance is necessary but not sufficient; a memory system must also leave behind user state that can be recovered later.
\end{insightbox}

\input{tables/tab_main_percat}

\subsection{Does Storing Evidence Make It Retrievable?}
Under \code{dump\_all}, \code{longctx\_full} is strongest ($B{=}0.624$), consistent with its role as a raw evidence-preservation baseline. Under \code{retrieve}, however, \code{amem} is strongest ($B{=}0.540$), while \code{longctx\_full} drops to $B{=}0.503$ and \code{mem0} to $B{=}0.473$. Thus, the best forensic archive is not necessarily the best operational memory. Raw turns preserve evidence, but they are not automatically organized as compact user-state claims that the read interface can surface. Evolved notes appear more compatible with operational retrieval, but their retrieve recovery remains only moderate, so this should not be read as solving the problem. \code{memt} shows the complementary failure of accumulation: it writes the largest store ($471$ items, $1015$ characters per item on average), yet does not obtain a retrieve-mode advantage ($B{=}0.465$), and its full-store dump exposes a budget-aware auditability problem because wholesale serialization exceeds the recovery-agent context. These patterns motivate write policies that are both update-aware and retrieval-aware, deciding when to store, merge, revise, or abstract traces into a compact recoverable user state.
\begin{insightbox}
\textbf{Finding 2.} Systems often keep the raw interaction but do not turn it into compact user state that retrieval can surface. Good memory should consolidate what it keeps into a form that is still usable when retrieved later; simply updating the store does not achieve this.
\end{insightbox}

\input{tables/tab_attrib}

\subsection{Why Is Stored Evidence Not Retrieved?}

Before diagnosing retrieval, we first validate the
disclosed-versus-undisclosed split used by the attribution pipeline.
Collapsing the human annotations to this binary decision yields
agreement in $71/86$ applicable cases ($82.6\%$, Cohen's
$\kappa{=}0.61$), and $22/30$ audited
\code{memory\_failure} cases contain human-confirmed disclosure.
The coarse diagnosis is also stable across attribution judges:
$91.6\%$--$96.6\%$ of failures are assigned to the combined
agent-or-memory side, with most variation confined to the finer
memory-versus-elicitation split
(Appendix~\ref{app:human_validation}).

Table~\ref{tab:attrib} separates failures caused by missing disclosure from failures caused by inaccessible stored evidence. In the canonical \code{dump\_all} rows for \code{amem}, \code{longctx\_full}, and \code{mem0}, task-design failures account for only $43$--$45$ dimensions, and agent-elicitation plus simulator-strictness failures account for only $44$--$55$ dimensions. Equivalently, $93.5\%$--$94.4\%$ of targets are either recovered or judged disclosed. Many failures therefore occur after the benchmark has created opportunities for disclosure and, in many cases, after relevant evidence appears in the interaction. Because attribution labels are assigned per access mode, a target recovered under \code{dump\_all} can re-enter attribution under \code{retrieve} when weak or cross-task evidence fails to surface; higher elicitation or simulator-strictness counts in retrieve rows should therefore not be read as changes to the task or simulator.

The remaining question is whether disclosed evidence was never preserved in the store, or whether it was preserved but not reachable through the normal read path. The \code{dump\_all}/\code{retrieve} comparison separates these cases. Table~\ref{tab:attrib} shows that disclosed-target recovery for the non-\code{memt} systems falls by roughly $8$--$18$ percentage points when access is restricted to top-$k$, and \code{memt} converges to the same range. Recoverable evidence can therefore exist in the written artifact while still failing to surface through the normal top-$k$ read interface. In this sense, \code{dump\_all} is a forensic probe of what the store contains, whereas \code{retrieve} measures whether that stored evidence is operationally accessible under the system's retrieval budget.
\begin{insightbox}
\textbf{Finding 3.} Evidence is often stored in memory yet cannot be reached through the system's normal retrieval. Storing evidence is not enough; the retrieval interface must actually bring it back when it is needed.
\end{insightbox}

\subsection{Which Memories Are Hardest to Recover?}
Table~\ref{tab:main_percat} shows that the category split reflects a contrast between directly stated interaction preferences and sparse, consequence-bound episodes, beyond any simple difficulty ranking. The same ordering appears in both \code{dump\_all} and \code{retrieve}, suggesting that the category gap is more than a retrieval artifact. \cat{Assistance Preference} is easiest across systems, because assistance dialogue naturally elicits explicit instructions about delivery style, such as tone, format, pacing, or level of detail. By contrast, \cat{Episodic Memory} is consistently the hardest category, with retrieve-mode recovery roughly a third to a half of the preference category, depending on the system.

These episodic targets are concrete, discrete, and time-anchored, often appearing only once and in narrow task-specific language. They are harder to recover mainly because the recovery target often depends on linking the event to its stated consequence, a demand that goes beyond quoting an isolated fact or preference; their rarity alone does not explain the gap. A writer can preserve the surface event while missing the ``when X happened, it led to Y'' relation that defines the episodic target. The consistent category gap therefore points to a binding problem, where systems must preserve the link between the event, the context in which it surfaced, and the consequence that defines the episodic target. This gap appears even though the user simulator is configured to volunteer relevant past events when the current scenario parallels its hidden episodic bank (Appendix~\ref{app:prompt_simulator}), so the deficit is unlikely to be explained solely by a silent simulator. \cat{Skill Memory}, \cat{Knowledge Memory}, and \cat{Self Model} fall between these poles, while \code{memt}'s retrieve results follow the same category shape; its uniformly low dump row reflects the out-of-context full-store auditability problem discussed above, not a category-specific weakness.
\begin{insightbox}
\textbf{Finding 4.} One-off, time-anchored past episodes are far harder to recover than stable, explicitly stated preferences. Episodic memory is more than logging events; a writer must keep the link between what happened, where it came up, and what it meant for the user.
\end{insightbox}

% \subsection{Case Study}
% Figure~\ref{fig:case-study-dimension-recovery} walks through a concrete $(\text{user}, \text{dimension})$ instance, where the task does not reveal the hidden bank entry but the user's follow-up turns expose evidence that the slot-fill evaluator partially recovers under both access modes. Appendix~\ref{app:gallery} provides additional audited cases with targets, tasks, diagnoses, and, where relevant, abridged transcripts or retrieved memories, per-system predictions, and judge reasoning. These cases show how the aggregate gaps arise in practice. Systems may preserve local interaction traces without abstracting the user-state claim they support; different writers may assign different abstraction levels to similar evidence; and stored evidence may fail to surface when abstract recovery queries do not match concrete, task-specific memories. Additional cases document boundary conditions such as slot-label polarity artifacts, localized wins from typed memory collections, and rare turn-budget failures, without changing the main diagnostic picture, in which current memory systems struggle to form compact user state from interaction traces and keep it accessible through the operational read interface. A small human audit (Appendix~\ref{app:human_validation}) confirms that both the simulator's \code{satisfied} signal and the Stage~0--2 attribution labels are broadly reliable for the analysis above.

\subsection{Case Study}
Figure~\ref{fig:case-study-dimension-recovery} walks through a concrete
$(\text{user}, \text{dimension})$ instance, where the task does not
reveal the hidden bank entry but the user's follow-up turns expose
evidence that the slot-fill evaluator partially recovers under both
access modes. Appendix~\ref{app:gallery} provides additional audited
cases with targets, tasks, diagnoses, and, where relevant, abridged
transcripts or retrieved memories, per-system predictions, and judge
reasoning. Together, these cases show how the aggregate gaps arise: systems may preserve local interaction traces without abstracting the user-state claims they support, assign different abstraction levels to similar
evidence, or fail to surface concrete, task-specific evidence using
abstract recovery queries. They also document boundary conditions,
including slot-label polarity artifacts, localized wins from typed
memory collections, and rare turn-budget failures, without changing
the main picture that current systems struggle to form compact,
operationally accessible user state. A small human audit (Appendix~\ref{app:human_validation}) confirms that both the simulator's \code{satisfied} signal and the Stage~0--2 attribution labels are broadly reliable for the analysis above.

\input{figures/fig_case_study_financial}

\subsection{Does Recovery Align with Later Personalization?}
\label{sec:personagym_validation}

As a complementary downstream check, we evaluate PersonaGym \citep{personagym}, giving the answering model only each system's native top-$5$ retrieved memories as user-specific context. Across $10{,}000$ answers for the four completed settings (\code{nomem}, \code{longctx\_full}, \code{amem}, and \code{mem0}), retrieve-mode recovery, PersonaScore, and the descriptive Memory-3
aggregate have the same numerical ordering (Table~\ref{tab:personagym_main}), and every memory-equipped
setting exceeds \code{nomem} on Memory-3. This provides complementary evidence that recoverable memory supports later persona-dependent behavior within the controlled setting;
Appendix~\ref{app:personagym} gives the full protocol, results, and
evaluation coverage.

\input{tables/tab_personagym_main}

%% file: tables/tab_main.tex
\begin{table}[htbp]
	\caption{
            Main results.
            $A$: task completion (\%);
            $D$: preference (/5);
            $C$: average turns;
            $B$: category-balanced recovery.
            \emph{Full store} is \code{dump\_all};
            \emph{Top-$k$} is \code{retrieve} with $k{=}5$.
            Footprint reports the number of items and mean characters per item.
            $^{\dagger}$\code{memt}'s overflowed full-store result is diagnostic
            rather than directly comparable.
            }
	\label{tab:main}
	\centering
	\small
	\setlength{\tabcolsep}{5pt}
	\begin{adjustbox}{max width=\linewidth}
		\begin{tabular}{@{}lccccccc@{}}
			\toprule
			System & Task ($A$,\,\%)\,$\uparrow$ & Preference ($D$)\,$\uparrow$ & Turns ($C$) & \multicolumn{2}{c}{Reconstruction ($B$)\,$\uparrow$} & \multicolumn{2}{c}{Footprint} \\
			\cmidrule(lr){5-6}\cmidrule(lr){7-8}
			 & & & & Full store & Top-$k$ & items ($N$) & chars \\
			\midrule
			\code{nomem}                     & ${99.935}$ & $4.585 \pm 0.175$          & $2.392 \pm 0.432$ & $0.000 \pm 0.000$          & --                          & 0   & 0     \\
			\code{amem}                      & ${99.935}$ & $4.640 \pm 0.157$          & $2.503 \pm 0.499$ & $0.611 \pm 0.062$          & $\mathbf{0.540 \pm 0.062}$ & 109 & 429   \\
			\code{longctx\_full}             & ${99.935}$ & $\mathbf{4.663 \pm 0.166}$ & $2.506 \pm 0.560$ & $\mathbf{0.624 \pm 0.067}$ & $0.503 \pm 0.075$          & 155 & 686   \\
			\code{mem0}                      & $99.871$          & $4.616 \pm 0.174$          & $2.460 \pm 0.465$ & $0.613 \pm 0.060$          & $0.473 \pm 0.079$          & 204 & 205   \\
			\code{memt}                      & ${99.935}$ & $4.631 \pm 0.161$          & $2.538 \pm 0.460$ & $0.130^{\dagger} \pm 0.251$ & $0.465 \pm 0.057$          & 471 & 1015  \\
			\bottomrule
		\end{tabular}
	\end{adjustbox}
\vspace{-8pt}
\end{table}

%% file: tables/tab_main_percat.tex
\begin{table}[tbp]
	\caption{Reconstruction score $B$ broken out by user-state category (mean $\pm$ std over $N{=}50$ users); $B \in [0,1]$, higher is better, and each cell is the mean reconstruction restricted to that category. \emph{Mode} is the read mode: \emph{Full store} reads the entire store (\code{dump\_all}), \emph{Top-$k$} the $k{=}5$ retrieved memories (\code{retrieve}). $^{\dagger}$The \code{memt} \emph{Full store} row is a diagnostic out-of-context full-store auditability outcome; it is not a comparable category-level reconstruction score.}
	\label{tab:main_percat}
	\centering
	\small
	\setlength{\tabcolsep}{4pt}
	\begin{adjustbox}{max width=\linewidth}
		\begin{tabular}{@{}llccccc@{}}
			\toprule
			System & Mode & Skill\,$\uparrow$ & Knowledge\,$\uparrow$ & Episodic\,$\uparrow$ & Self-model\,$\uparrow$ & Preference\,$\uparrow$ \\
			\midrule
			\code{amem}          & Full store & $0.636 \pm 0.089$          & $0.650 \pm 0.071$          & $0.416 \pm 0.126$          & $0.580 \pm 0.135$          & $\mathbf{0.773 \pm 0.060}$ \\
			\code{amem}          & Top-$k$    & $\mathbf{0.567 \pm 0.102}$ & $\mathbf{0.544 \pm 0.105}$ & $\mathbf{0.351 \pm 0.144}$ & $\mathbf{0.516 \pm 0.158}$ & $0.721 \pm 0.089$          \\
			\midrule
			\code{longctx\_full} & Full store & $0.645 \pm 0.095$ & $\mathbf{0.657 \pm 0.069}$ & $\mathbf{0.451 \pm 0.129}$ & $\mathbf{0.609 \pm 0.143}$ & $0.759 \pm 0.066$ \\
			\code{longctx\_full} & Top-$k$    & $0.534 \pm 0.118$ & $0.504 \pm 0.109$          & $0.297 \pm 0.133$          & $0.444 \pm 0.164$          & $\mathbf{0.736 \pm 0.082}$ \\
			\midrule
			\code{mem0}          & Full store & $\mathbf{0.651 \pm 0.086}$ & $0.639 \pm 0.068$ & $0.423 \pm 0.125$ & $0.588 \pm 0.142$ & $0.765 \pm 0.092$          \\
			\code{mem0}          & Top-$k$    & $0.516 \pm 0.124$          & $0.439 \pm 0.151$ & $0.242 \pm 0.125$ & $0.431 \pm 0.166$ & $\mathbf{0.736 \pm 0.088}$ \\
			\midrule
			\code{memt}          & Full store$^{\dagger}$ & $0.129 \pm 0.253$ & $0.137 \pm 0.263$ & $0.089 \pm 0.180$ & $0.133 \pm 0.262$ & $0.164 \pm 0.314$ \\
			\code{memt}          & Top-$k$    & $0.535 \pm 0.114$ & $0.411 \pm 0.114$ & $0.229 \pm 0.117$ & $0.431 \pm 0.151$ & $0.720 \pm 0.086$ \\
			\bottomrule
		\end{tabular}
	\end{adjustbox}
\vspace{-8pt}
\end{table}

%% file: tables/tab_attrib.tex
\begin{table}[hbp]
	\caption{
            Failure attribution over $1{,}550$ targets per system.
            Rows partition targets into recovered (per-target recovery score $\geq 0.75$),
            post-disclosure recovery failure (\emph{Memory}),
            task-design failure (\emph{Task}),
            agent-elicitation failure (\emph{Agent}),
            and simulator non-disclosure (\emph{User}).
            Recovery rate is
            $\text{Recovered}/(\text{Recovered}+\text{Memory})$.
            Unsuffixed rows use \code{dump\_all};
            top-$k$ rows use \code{retrieve}.
            $^{\dagger}$\code{memt}'s overflowed full-store row is diagnostic.
            }
	\label{tab:attrib}
	\centering
	\footnotesize
	\setlength{\tabcolsep}{6pt}
\renewcommand{\arraystretch}{0.95}
	\begin{adjustbox}{max width=\linewidth}
		\begin{tabular}{@{}lrrrrrr@{}}
			\toprule
			System & Recovered\,$\uparrow$ & Memory\,$\downarrow$ & Task\,$\downarrow$ & Agent\,$\downarrow$ & User\,$\downarrow$ & Recovery rate\,$\uparrow$ \\
			\midrule
			\code{nomem}                 & 0   & 1410 & 73 & 62  & 5  & $0.00\%$           \\
			\code{amem}                  & 950 & 500  & 45 & 51  & 4  & $65.52\%$          \\
			\code{amem} (top-$k$)          & 769 & 578  & 49 & 114 & 40 & $\mathbf{57.09\%}$ \\
			\code{longctx\_full}         & 982 & 481  & 43 & 41  & 3  & $\mathbf{67.12\%}$ \\
			\code{longctx\_full} (top-$k$) & 682 & 666  & 56 & 107 & 39 & $50.59\%$          \\
			\code{mem0}                  & 961 & 494  & 44 & 46  & 5  & $66.05\%$          \\
			\code{mem0} (top-$k$)          & 628 & 687  & 55 & 132 & 48 & $47.76\%$          \\
			\code{memt}$^{\dagger}$      & 197 & 1219 & 69 & 60  & 5  & $13.91\%$          \\
			\code{memt} (top-$k$)          & 624 & 706  & 60 & 118 & 42 & $46.92\%$          \\
			\bottomrule
		\end{tabular}
	\end{adjustbox}
\vspace{-8pt}
\end{table}

%% file: figures/fig_case_study_financial.tex
\newcolumntype{L}[1]{>{\raggedright\arraybackslash}p{#1}}
\newcolumntype{Y}{>{\raggedright\arraybackslash}X}

\begin{figure}[t]
\centering
\scriptsize
\begin{tcolorbox}[
    width=1.0\linewidth,
    colback=gray!3,
    colframe=gray!50,
    boxrule=0.55pt,
    arc=1.2mm,
    boxsep=1pt,
    before skip=0pt,
    after skip=0pt,
    left=2pt,
    right=2pt,
    top=1.5pt,
    bottom=1.5pt,
    title=\textbf{Case Study: Recovering a Hidden User Dimension from Interaction},
    fonttitle=\bfseries,
    coltitle=black,
    colbacktitle=gray!15
]

\textbf{Hidden ground-truth dimension:}
\texttt{financial\_literacy\_knowledge}
\hfill
\textit{``Understands money as risk, fees, and avoidable loss.''}

\vspace{0.05em}
\textbf{Task:}
The agent helps the user decide whether to pay off a car loan early or continue regular payments, while identifying fees and hidden payoff costs.

\vspace{0.12em}

\renewcommand{\arraystretch}{0.94}
\setlength{\tabcolsep}{2.5pt}
\begin{tabularx}{\linewidth}{L{0.04\linewidth} Y L{0.29\linewidth}}
\toprule
\textbf{Turn} & \textbf{User evidence} & \textbf{Exposed signal} \\
\midrule

T1 &
``I want the actual decision points laid out in order \ldots{} how to compare the loan rate against what the cash could earn somewhere \ev{safe} \ldots{} whether paying it off early can \ev{hurt liquidity enough to matter} \ldots{} sending the \ev{wrong amount or missing daily interest}.'' &
Frames payoff as a risk, liquidity, and avoidable-loss decision. \\

T2 &
``If the loan APR is above the safest return you can get \ev{after tax}, and you'll still have a real emergency fund left, pay it off \ldots{} any \ev{prepayment penalty} gets added into the math immediately \ldots{} not some \ev{fantasy return}.'' &
Uses conservative financial comparison: after-tax safe return, explicit fees, and non-speculative opportunity cost. \\

T3 &
``If paying it off would leave the emergency fund below a real floor, stop there and keep paying \ldots{} something like \ev{3--6 months of bare expenses} \ldots{} quote expires, interest keeps running \ldots{} underpay by even a small amount, account stays open.'' &
Applies a hard liquidity floor and treats payoff-process mistakes as concrete financial failure points. \\

T4 &
``If the loan APR is higher than what I can get in savings or money market \ev{after tax} by at least a little margin, PAY OFF \ldots{} if the gap is tiny, KEEP PAYING and don't \ev{burn liquidity for crumbs}.'' &
Shows cautious, loss-avoidant reasoning: small interest savings should not override liquidity protection. \\

T5 &
``If the gap is barely there, I'd lean KEEP PAYING and \ev{not burn liquidity for crumbs}.'' &
Confirms that the user prioritizes avoiding unnecessary liquidity risk over marginal payoff gains. \\

\bottomrule
\end{tabularx}

\vspace{0.15em}

\begin{tcolorbox}[
    colback=green!4,
    colframe=green!40!black,
    boxrule=0.35pt,
    arc=0.9mm,
    boxsep=1pt,
    before skip=2pt,
    after skip=0pt,
    left=2pt,
    right=2pt,
    top=1.5pt,
    bottom=1.5pt
]
\textbf{Final predictions from memory.}
\begin{itemize}[leftmargin=1.0em, itemsep=0pt, topsep=0pt, parsep=0pt, partopsep=0pt]
    \item \texttt{dump\_all}: \emph{``understands practical personal finance and billing disputes''}
    \item \texttt{retrieve}: \emph{``understands debt payoff tradeoffs and dispute basics, wants precise rules''}
\end{itemize}
\end{tcolorbox}

\end{tcolorbox}
\caption{
The target is hidden from the agent; highlighted phrases expose diagnostic evidence.
}
\label{fig:case-study-dimension-recovery}
\end{figure}

%% file: tables/tab_personagym_main.tex
\begin{table}[hbp]
\centering
\footnotesize
\caption{Downstream PersonaGym validation ($50$ users; $10{,}000$ answers across four settings). Memory-3 is the descriptive mean of Linguistic Habits, Persona Consistency, and Action Justification. Bold marks the best value in each column, including ties.}
\label{tab:personagym_main}
\setlength{\tabcolsep}{3pt}
\renewcommand{\arraystretch}{1.0}
\begin{adjustbox}{max width=\linewidth}
\begin{tabular}{@{}lcccccccc@{}}
\toprule
System & \shortstack{Retrieve-mode\\recovery $B$} & \shortstack{Expected\\Action} & Toxicity & \shortstack{Linguistic\\Habits} & \shortstack{Persona\\Consistency} & \shortstack{Action\\Justification} & PersonaScore & \shortstack{Memory-3\\mean} \\
\midrule
\code{nomem} & 0.000 & 4.25 & \textbf{5.00} & 2.72 & 3.88 & 3.57 & 3.882 & 3.389 \\
\code{longctx\_full} & 0.503 & 4.33 & 4.97 & \textbf{3.29} & 4.15 & \textbf{3.77} & 4.101 & 3.734 \\
\code{amem} & \textbf{0.540} & \textbf{4.37} & 4.97 & \textbf{3.29} & \textbf{4.23} & \textbf{3.77} & \textbf{4.128} & \textbf{3.766} \\
\code{mem0} & 0.473 & 4.36 & 4.98 & 3.20 & 4.20 & 3.75 & 4.099 & 3.716 \\
\bottomrule
\end{tabular}
\end{adjustbox}
\end{table}

%% file: sections/6_conclusion.tex
\section{Conclusion}
\label{sec:conclusion}
\bench{} reframes long-term agent memory as an auditable post-interaction artifact: after ordinary assistance, what hidden user state can be reconstructed from the memory the agent leaves behind? Across five memory settings, near-saturated task completion coexists with only moderate full-store recovery and substantially lower top-$k$ recovery, with system rankings changing between access modes. Successful assistance therefore does not entail recoverable memory. These results show that evidence preservation, retrieval-aware state formation, operational accessibility, and budget-aware consolidation form distinct stages of the long-term memory lifecycle, which a single memory-quality axis cannot capture. Within its controlled synthetic setting, \bench{} suggests that reliable memory agents need more than larger stores or better short-term responses; they need memory artifacts that stay recoverable after interaction.

%% file: sections/7_acknowledge.tex
\clearpage
\section{Acknowledgments}
\label{sec:Acknowledgments}

This work used DeltaAI and Delta CPU computing resources at the National Center for Supercomputing Applications (NCSA) through allocation CIS260387 from the Advanced Cyberinfrastructure Coordination Ecosystem: Services \& Support (ACCESS) program, which is supported by U.S. National Science Foundation grants \#2138259, \#2138286, \#2138307, \#2137603, and \#2138296~\cite{Boerner2023ACCESS}.

%% file: appendix/limitations.tex
\section{Limitations}
\label{app:limitations}

\paragraph{Synthetic-user and scaffold scope.}
\bench{} uses synthetic users and generated tasks to make memory recovery measurable under known ground truth. This setting reflects a common assistant regime: users work on immediate tasks, while longer-term signals surface through constraints, corrections, preferences, prior experiences, and repeated behavior. Synthetic profiles let us control the hidden user state, avoid prompt leakage, and audit whether disclosed evidence becomes recoverable memory. The current benchmark uses only U.S.-based DeepPersona profiles to keep language and cultural assumptions comparable during scoring and audit. As a result, \bench{} supports controlled comparison of memory mechanisms, but should not be read as covering all real-user behavior, especially multilingual, cross-cultural, highly privacy-protective, or rapidly shifting user settings. The broader experimental scaffold is likewise intentionally controlled: the data-generation pipeline, user simulator, assistant scaffold, dialogue protocol, and judge-based scoring are structured to make memory formation measurable and failures attributable; they do not attempt to fully simulate open-ended human--assistant use. This choice prioritizes auditability and controlled comparison, while leaving richer models of user behavior, assistant adaptation, dialogue dynamics, and human evaluation to future work. A second design choice biases recovery in the opposite direction: the bank generator is instructed to select only pool dimensions on which the user has a meaningful, non-generic stance (Appendix~\ref{app:prompt_bank}), so the 31 dimensions per user are concentrated on user-strong axes. Recovery numbers therefore reflect performance on dimensions designed to be disclosable, and absolute scores should be read as upper bounds relative to a uniform sample over the dimension pool.

\paragraph{Scoring and attribution scope.}
Recovery scores and attribution labels are defined by the rubric-based judging protocol used in \bench{}. This follows prior LLM-as-a-judge evaluation work that uses model judges for scalable assessment, while acknowledging that judge reliability must itself be audited and interpreted cautiously \citep{zheng2023judgingllmasajudgemtbenchchatbot,tan2025judgebenchbenchmarkevaluatingllmbased}. The recovery judge is shown the dimension, its explanation, the predicted label, and the ground-truth label simultaneously, without blinding (Appendix~\ref{app:prompt_judge}); this is standard in scalable LLM-judge protocols but means scores are sensitive to surface-form choices in the predicted label. Some slot names are also sensitive to formulation, especially directional axes whose polarity can be verbalized in more than one way; Case~3 in Appendix~\ref{app:gallery} is an audited example of how this can flip the judge's verdict despite consistent underlying behavior. For such cases, a low score may not uniquely identify whether the error occurred during memory writing, slot filling, or judge interpretation. To support audit, we release prompts, slot-fill outputs, judge scores, rationales, and attribution records so that individual decisions can be inspected. The resulting scores support comparisons under the stated rubric, but should not be over-interpreted as human gold labels.

\paragraph{Task difficulty and coverage.}
\bench{} uses everyday assistance tasks, with no adversarial or highly specialized cases. This choice keeps the episode objective realistic and makes task fit easy to satisfy: in our runs, task completion nearly saturates across systems. The benchmark should therefore not be read as a stress test of general task-solving ability. Its focus is narrower: even when the immediate task is successfully completed, the interaction may still fail to leave behind recoverable user state. Future variants could add more specialized domains, longer projects, or harder multi-step tasks to test whether memory failures change when task success itself is less saturated.

\paragraph{User-conditioned assistance skills not directly evaluated.}
The attribution results point to a memory target beyond user facts and preferences. Some failures occur before writing, when the task could invite the target but the assistant does not elicit enough evidence. Across repeated conversations, an agent could plausibly learn how to work with this particular user: useful unknowns, productive prompts, and feedback trajectories. These are skill-like memories in form, but user-conditioned and interaction-centered, keyed to the particular user across sessions. \bench{} does not directly evaluate this memory type, but separating task-design, elicitation, writing, retrieval, and budget failures makes the need visible. We leave a deeper investigation to future work.

%% file: appendix/broader_impact.tex
\section{Broader Impact}
\label{app:broader_impact}

\paragraph{Implications for memory-system design.}
The results suggest that long-term memory should be evaluated as an auditable post-interaction artifact, a view that subsumes its roles as a write--retrieve buffer and a source of downstream task gains. At the same time, ``dynamic memory'' alone does not name the missing capability. Existing systems already include dynamic note linking and memory evolution \citep{xu2025amemagenticmemoryllm}, conversational memory consolidation and update operations \citep{chhikara2025mem0buildingproductionreadyai}, learned memory-operation policies \citep{yue2026memtdensifyingrewardslonghorizon}, and reflection mechanisms that synthesize interaction traces into higher-level inferences \citep{park2023generativeagentsinteractivesimulacra,shinn2023reflexionlanguageagentsverbal}. What remains missing is \emph{recovery-aligned state formation}: memory dynamics that leave behind compact, updated, and retrievable user-state claims after interaction, a property distinct from improving downstream responses or maintaining a plausible memory store. This points to three concrete directions. First, memory operations should be evaluated by what user-state artifact they produce, which the add, update, delete, link, and retrieve view does not capture. Second, retrieval should be evaluated together with written-store recovery when the evaluation setting makes this diagnostically possible, since the bounded full-store probe in \bench{} is a forensic diagnostic that separates evidence preservation from operational accessibility. Third, memory systems need budget-aware consolidation: a larger and more structured store can become less auditable without producing a clear retrieve-mode advantage.

\paragraph{Learned memory-management policies are promising but incomplete.}
The \code{memt} comparison should not be read as evidence against learned or RL-based memory management. The direction is natural: write, update, delete, and retrieve decisions are sequential, context-dependent, and coupled to future utility. However, in our setting, a trained operation policy does not automatically translate into recoverable user state. \code{memt} writes the largest store and uses a learned memory-operation policy, yet it does not obtain a retrieve-mode advantage, and its full-store serialization creates an auditability problem. This suggests that current formulations may improve procedural operation selection without yet learning robust abstraction, consolidation, conflict handling, or retrieval-aware organization. Future learned memory managers may need rewards and operation spaces that directly optimize recoverable, compact, and accessible user-state formation, a target that the choice of memory operations alone does not reach.

\paragraph{Privacy, profiling, and consent.}
\bench{} is intended to make long-term memory systems more auditable by measuring what user state remains recoverable after interaction. This can support safer memory-system design by exposing over-accumulation, retrieval failures, and auditability problems. At the same time, stronger user-memory systems can increase privacy, profiling, and unwanted personalization risks if deployed on real users without consent, data minimization, deletion controls, or inspection mechanisms. The released benchmark therefore uses synthetic users and should not be treated as a deployment-ready personalization dataset or a psychometric model of real users.

%% file: appendix/benchmark_card.tex
\section{Benchmark Card and Release Artifacts}
\label{app:benchmark_card}
\input{tables/tab_token_cost}
\input{tables/tab_data_construction_token_cost}
\input{tables/tab_release_artifacts}
\paragraph{Intended use.} \bench{} is intended for controlled comparison of
LLM-agent memory systems on post-interaction structured user-state recovery. It
is suitable for ablating memory-system design choices (write policy, retrieval
interface, footprint trade-off) under fixed task and simulator conditions. It is
\emph{not} a stress test of general assistant capability, a personalization
benchmark, or a model of real human users.
\paragraph{Out-of-scope use.} The structured banks are taxonomy-anchored
synthetic targets used to make recoverable user state observable; they are not
intended as psychometric assessments, recommendations for personalization
deployment, or evidence of system safety in adversarial settings. The U.S.-only
DeepPersona profile pool was chosen for language and cultural coherence during
audit, and is not representative of multilingual or cross-cultural assistant
populations.
\paragraph{Reproducibility.} Benchmark instances (banks, tasks) and full
evaluation traces are released as JSON. Aggregate metrics in
Tables~\ref{tab:main}--\ref{tab:attrib} are recomputable from the per-user
records using the released analysis scripts. All judge / oracle / disclosure /
slot-fill calls run at \code{temperature=0.0} to minimize sampling
variance; outputs are typically stable across reruns but not
bit-deterministic, because API-side numerical precision and inference
infrastructure introduce small non-deterministic variation even at
\code{temperature=0.0}. We mitigate per-call variability by reporting
aggregate scores over $50$ users $\times$ $31$ dims and by releasing every
judge prediction and rationale so individual decisions remain auditable.
\paragraph{Token cost (for reference).}
Table~\ref{tab:token_cost} covers \code{gpt-5.4-mini} API-side calls, including
agent replies where applicable, simulator calls, preference judging, write-side memory calls,
slot-filling, and recovery / attribution judging. \code{memt}'s memory-write and
ReAct-retrieval policy uses Mem-T-4B served
locally via vLLM on a single NVIDIA A100 GPU and is not included in the
\code{gpt-5.4-mini} totals below. The table reports the cumulative
\code{gpt-5.4-mini} token usage logged across all $50$-user benchmark sessions for each
system, excluding the one-time offline data-construction cost reported
separately in Table~\ref{tab:data_construction_token_cost}. The relative
ordering across systems is intuitive: \code{longctx\_full} is highest because
its \code{dump\_all} slot-fill receives the full verbatim transcript store
(median store size: $150$ items and $625$ characters per item per user);
\code{mem0} is the lightest memory-equipped system because its atomic-fact
writes are short and its store is the most compact; \code{memt}'s \code{gpt-5.4-mini}
share is moderate because the bulk of its memory operations runs on Mem-T-4B
locally. We emphasize that the absolute numbers in Table~\ref{tab:token_cost}
are not exact: they accumulate benchmark-iteration and server-crash-recovery
sessions, so they are upper bounds on a clean single-pass cost and the
per-system inflation factor is not uniform. They are included for orientation only
and should not ground fine-grained cost comparison.

%% file: tables/tab_token_cost.tex
\begin{table}[t]
    \caption{Cumulative \texttt{gpt-5.4-mini} token usage across the $50$-user benchmark, in millions of tokens, \emph{for reference only}. These are non-auditable aggregate run-log records and are not part of the recomputable release artifacts. The per-system rows cover episode-loop calls (agent reply, simulator, preference judge, write-side LLM calls) and recovery scoring where applicable (\emph{dump\_all} + \emph{retrieve} slot-fill and recovery judge); the attribution row (Stage~0/1/2 judges) is reported separately because Stage~0 is cached per \((user, dimension)\) and shared across all five systems. Excludes \code{memt}'s local Mem-T-4B inference. The numbers are upper bounds because they accumulate token usage across server-crash recovery and benchmark-iteration sessions; a clean single-pass run would consume meaningfully less and the per-system inflation factor is not uniform, so these figures should not be used for fine-grained cost comparisons across systems.}
    \label{tab:token_cost}
    \centering
    \small
    \setlength{\tabcolsep}{8pt}
    \begin{adjustbox}{max width=\linewidth}
        \begin{tabular}{@{}lrrr@{}}
            \toprule
            component & prompt (M) & completion (M) & total (M) \\
            \midrule
            \code{nomem}                     &  61.05 & 6.07 &  67.12 \\
            \code{amem}                      & 126.82 & 9.04 & 135.85 \\
            \code{longctx\_full}             & 238.12 & 4.82 & 242.94 \\
            \code{mem0}                      &  88.16 & 3.99 &  92.15 \\
            \code{memt}$^{\dagger}$          & 120.91 & 3.77 & 124.68 \\
            \midrule
            Subtotal (5 systems)             & 635.06 & 27.69 & 662.74 \\
            Attribution (shared, Stage~0/1/2 across all runs) & 25.47 & 1.70 & 27.17 \\
            \midrule
            \textbf{Grand total}             & \textbf{660.53} & \textbf{29.39} & \textbf{689.91} \\
            \bottomrule
        \end{tabular}
    \end{adjustbox}
    \\[2pt]
    \footnotesize
    $^{\dagger}$\,\texttt{gpt-5.4-mini} side only. \code{memt}'s memory-operation policy runs on Mem-T-4B served locally via vLLM on a single NVIDIA A100 GPU and is not included in the totals above. Per-call API usage logs are not part of the public release; the totals here are reported as non-auditable aggregate figures only.
\end{table}

%% file: tables/tab_data_construction_token_cost.tex
\begin{table}[t]
    \caption{One-time offline token usage for constructing the released $50$-user benchmark pool, in millions of tokens, reported as a non-auditable aggregate run-log record for reference only. This cost covers persona-profile generation, hidden-bank generation, task generation, anti-fishing critic calls, and editor revisions. It is separate from the per-system benchmark-run costs in Table~\ref{tab:token_cost}.}
    \label{tab:data_construction_token_cost}
    \centering
    \small
    \setlength{\tabcolsep}{7pt}
    \begin{adjustbox}{max width=\linewidth}
        \begin{tabular}{@{}lrrrr@{}}
            \toprule
            stage & calls & prompt (M) & completion (M) & total (M) \\
            \midrule
            Data construction for $50$ users & 1,721 & 17.04 & 0.17 & 17.21 \\
            \bottomrule
        \end{tabular}
    \end{adjustbox}
    \\[2pt]
    \footnotesize
    All calls use \code{gpt-5.4-mini}.
\end{table}

%% file: tables/tab_release_artifacts.tex
\begin{table}[t]
	\caption{Released artifact package.}
	\label{tab:release_artifacts}
	\centering
	\small
	\setlength{\tabcolsep}{4pt}
	\begin{adjustbox}{max width=\linewidth}
		\begin{tabular}{@{}p{0.23\linewidth}p{0.70\linewidth}@{}}
			\toprule
			artifact & description \\
			\midrule
			Generation configs & Bank schema (5 categories, 7/7/7/5/5 dims per user, 31 dims total), random seeds, $T{=}31$ tasks per user, 25-turn cap, scoring modes (\emph{dump\_all}, \emph{retrieve}). LLM identifiers used (judge, slot-fill, simulator, agent, write-side) are documented in the release README; per-call API logs are not part of the public release. \\
			Bank records & 50 user banks ($1{,}550$ ground-truth dims) with \texttt{category}, \texttt{dimension}, \texttt{short} target phrase, \texttt{explanation}, and source-taxonomy mapping. Released as \texttt{user\_memory\_banks\_pooled\_final.json}. \\
			Task records & $1{,}550$ generated tasks, target dimension, blind anti-fishing critic verdict (\emph{neutral} / \emph{fishing}), and accepted-after-revision metadata. Released under \texttt{benchmark\_data/CustomTasksPooledFinal/}. \\
			Dialogue transcripts & Full per-turn transcripts under \path{history/<run_id>/user_*/episode_*.json}, including agent reply, simulator feedback type and text, preference judge output, and \texttt{end\_reason}. \\
			Memory artifacts & Per-user memory store at end of $T$-task interaction (\texttt{memory/<run\_id>/user\_*/memories.json}); for \emph{retrieve} mode, the actual top-$k$ memories returned by \texttt{agent.search(query, k=5)} are recorded inside each \texttt{recon\_judge} entry. \\
			Scoring artifacts & Slot-fill prompts and predictions, recovery-judge scores and rationales, per-category aggregates, and the merged 50-user reports under \texttt{output/<run\_id>/}. \\
			Attribution artifacts & Stage~0 task-design oracle outputs (cached per \((user, dimension)\) and shared across runs), Stage~1 disclosure-judge outputs, Stage~2 subclass labels, and the final attribution category per dim under \texttt{output/<run\_id>/attribution/}. \\
			Pipeline code & Modular Python implementation of the full pipeline: \texttt{runner.py} (episode-loop orchestrator and unified scoring entry point), \texttt{simulation.py} (persona-conditioned user simulator and episode controller), \texttt{scorer.py} (slot-fill and recovery judge), \texttt{failure\_attribution.py} (Stage~0/1/2 attribution), \texttt{task\_generator.py} (per-dimension task generation with the anti-fishing critic and editor), \texttt{Deeppersona/generate\_memory\_bank\_pooled.py} (taxonomy-anchored bank generation), and the per-system agents under \texttt{agents/} (\code{nomem}, \code{longctx\_full}, \code{amem}, \code{mem0}, \code{memt}). \\
			Analysis scripts & Stand-alone scripts for re-aggregating Tables~\ref{tab:main}, \ref{tab:main_percat}, \ref{tab:attrib} from the per-user records: \texttt{merge\_50u\_reports.py} and \texttt{failure\_attribution.py}. \\
			\bottomrule
		\end{tabular}
	\end{adjustbox}
\end{table}

%% file: appendix/generation.tex
\section{Generation Pipeline Details}
\label{app:generation}
For each user, generation runs in three stages: (1) sample a DeepPersona base
profile \citep{wang2025deeppersonagenerativeenginescaling}, on which all
subsequent generation steps are conditioned; (2) for each category, the bank
generator picks dims from a fixed taxonomy-anchored pool and fills each
with a user-specific \code{short} target phrase plus a 2--4 sentence
\code{explanation} grounded in the persona; (3) the task generator produces
one assistance task per $(\text{user}, \text{dimension})$, routed through a blind
anti-fishing critic that rejects or revises tasks whose answers are recoverable
from the task statement alone. The bank generator validates that all chosen dim
names come verbatim from the pool and retries on failure; the task generator
retries until either the critic accepts the task or a retry budget is reached.
\input{tables/tab_bank_categories}

%% file: tables/tab_bank_categories.tex
\begin{table}[H]
	\caption{Structured bank categories used by the released $N{=}50$ pool. Slot counts are per user; the total of $31$ dims per user $\times\,50$ users $=\,1{,}550$ recovery targets matches the entries reported in Tables~\ref{tab:main}--\ref{tab:attrib}.}
	\label{tab:bank_categories}
	\centering
	\small
	\setlength{\tabcolsep}{4pt}
	\begin{adjustbox}{max width=\linewidth}
		\begin{tabular}{@{}p{0.22\linewidth}p{0.10\linewidth}p{0.24\linewidth}p{0.36\linewidth}@{}}
			\toprule
			category & slots / user & source & role in benchmark \\
			\midrule
			\cat{Skill Memory} & 7 & O*NET work-style and ability descriptors \citep{Online2026-ci} & Recoverable ability or task-style trait, e.g.\ ``negotiates firmly'', ``troubleshoots household failures''. \\
			\cat{Knowledge Memory} & 7 & O*NET domain-knowledge descriptors \citep{Online2026-ci} & Domain familiarity or knowledge trace, e.g.\ ``functional local sense of roads and region'', ``everyday math sense, not advanced science''. \\
			\cat{Episodic Memory} & 7 & Conway autobiographical-memory taxonomy and McAdams life-story constructs \citep{CONWAY2005594,McAdams1997TheSW} & Sparse, time-anchored autobiographical event with a recoverable consequence, e.g.\ ``moment of clarity'', ``moral dilemma'', ``reminiscence-bump event''. \\
			\cat{Self Model} & 5 & HEXACO facets \citep{Ashton2007-vl,Lee2018-jf} remapped to user-level traits & Abstract self-view, e.g.\ ``quietly fair, by-the-book'', ``anchors identity in independence and control''. \\
			\cat{Assistance Preference} & 5 & Operational interaction-style descriptors & How the user prefers assistance, e.g.\ verbosity, jargon tolerance, pushback tolerance, format preference. \\
			\midrule
			Total per user & 31 & -- & $50$ users $\times\,31$ dims $=\,1{,}550$ recovery targets per memory system. \\
			\bottomrule
		\end{tabular}
	\end{adjustbox}
\end{table}

%% file: appendix/prompts.tex
\section{Prompt Templates}
\label{app:prompts}

The prompts below are reproduced from the released codebase with print-only typography normalization: box-drawing characters and curly quotes have been ASCII-normalized so they render cleanly, while the prompt logic is unchanged. Curly-brace placeholders (e.g.\ \verb|{dimension}|, \verb|{transcript}|) are filled in at call time by the corresponding Python module.

\subsection{Bank generation prompt}
\label{app:prompt_bank}
This is the system prompt for bank generation in \path{Deeppersona/generate_memory_bank_pooled.py}.
\begin{promptbox}
\begin{Verbatim}[fontsize=\scriptsize,breaklines=true,breakanywhere=true,breakautoindent=false,breakindent=0pt,breaksymbolleft={},breaksymbolsepleft=0pt]
You are generating one category of a user's hidden memory bank by PICKING dims
from a fixed pool and filling each with a user-specific short + explanation.

HARD RULES:
- Pick EXACTLY the requested count of dim names from the provided pool.
  You MUST NOT invent, rename, misspell, or merge pool entries. Names must be
  copied verbatim from the pool.
- Only select pool dims where THIS user has a meaningful, non-generic stance
  on that axis. If the user has no clear stance, skip it and pick a better fit
  from the pool.
- Prefer LATERAL DIVERSITY -- avoid picking multiple dims that all point at the
  same underlying trait.
- "short": <=15 words, user-specific label. Concrete, directly expressing the
  user's stance on the axis.
- "explanation": 2-4 sentences. Grounded in the persona, concrete, specific to
  this user. Gaps, misconceptions, and weaknesses matter as much as strengths.
- Stay strictly within the stated category boundary.

Output JSON only -- no markdown fences, no commentary:
{
  "dims": [
    {"name": "<pool dim name, verbatim>", "short": "...", "explanation": "..."},
    ...
  ]
}
\end{Verbatim}
\end{promptbox}

\subsection{Task generation prompt}
\label{app:prompt_task}
This is the system prompt for the per-dimension task generator in \texttt{task\_generator.py}.
\begin{promptbox}
\begin{Verbatim}[fontsize=\scriptsize,breaklines=true,breakanywhere=true,breakautoindent=false,breakindent=0pt,breaksymbolleft={},breaksymbolsepleft=0pt]
You design evaluation tasks for a memory-benchmark study. Each task is the FIRST message
the user sends to an AI assistant. Its job is to create a realistic help context in which
a specific, pre-known aspect of the user will organically surface through the user's reply,
NOT through the task text itself.

HARD RULES:
1. The task MUST be a help request / planning request / recommendation request / problem-solving
   request.
2. The task MUST NOT directly probe the user ("what are you good at", "how do you like to be
   helped", "tell me about a time you...") -- those are silly exposure prompts. Zero tolerance.
3. The task MUST NOT copy phrasing from the target memory entry. The entry content is what we
   hope the user will reveal in THEIR REPLY; leaking it in the task defeats the purpose.
4. The task must be plausible first-contact text: 1-3 sentences, in a voice the user would use.
5. The task must be plausibly addressable by a generalist AI assistant (no insider info,
   no physical-world actions required from the assistant).
6. The task must be plausible given the user's actual profile (values, life situation, interests).
7. The task MUST actually expose the target memory entry through the user's natural reply.
   Exposure is the only reason this task exists -- if a framing looks novel but would not
   draw out the target entry, it is useless. Reject it and redo.
8. Scene diversity: prefer a scene/domain that the prior tasks for this user have NOT already
   used (typical scenes: household, health, finance, travel, shopping, social, tech, hobbies,
   food, work/volunteering, transportation). BUT rule 7 wins: if genuine exposure of THIS
   entry demands a scene that was already used, reuse it rather than picking a mismatched
   scene that weakens exposure.
9. ANTI-FISHING (situation vs trait): SITUATION-level framing is fine -- describe a neutral
   circumstance the user plausibly faces. TRAIT-level framing is fishing -- do NOT echo the
   dimension name, short, explanation, or a paraphrase of the trait, and do NOT directly
   solicit self-description. A third party reading ONLY the task text must not be able to
   guess which trait is being probed; exposure must come from the USER'S REPLY given their
   profile, not from the task's wording.
   - BAD (echoes dimension phrasing): "keep better track of service calls" when dimension
     is 'recordkeeping_precision' -- "tracking" is the trait itself.
   - BAD (paraphrases the trait): "a fallback plan that doesn't get messy" when dimension
     is 'low_disruption_adaptation' -- "doesn't get messy" restates "low disruption".
   - BAD (solicits self-description): "help me write a personal statement that sounds like
     me" -- directly asks the user to describe themselves.
   - BAD (names the trait): "how much detail is actually needed" when dimension is
     'privacy_orientation' -- names the privacy decision explicitly.
   - GOOD (situation setup, not trait echo): "I've got a problem bigger than a simple fix,
     how do I handle it right?" for 'problem_escalation' -- the situation is neutral; the
     user's handling style (escalate / DIY / call pros / document) emerges in THEIR reply.
   - GOOD (open-ended): "a couple of repairs lined up this month -- what should I think
     about before I call anyone?" -- records-habits, escalation, voice, preferences all
     surface in the reply, nothing in the task telegraphs any single trait.

OUTPUT: strict JSON, no markdown fence, no commentary.
\end{Verbatim}
\end{promptbox}

\subsection{Anti-fishing critic prompt}
\label{app:prompt_antifishing}
This is the system prompt for the blind anti-fishing critic; tasks flagged \emph{fishing} are passed through the editor below.
\begin{promptbox}
\begin{Verbatim}[fontsize=\scriptsize,breaklines=true,breakanywhere=true,breakautoindent=false,breakindent=0pt,breaksymbolleft={},breaksymbolsepleft=0pt]
You are a BLIND anti-fishing auditor for evaluation tasks. You see ONLY the task text and
the memory category it claims to probe -- NOT the user, NOT the target entry.

Your job: given only the task, guess the most likely trait / stance / habit the task
telegraphs about its sender. Then classify the task:

- "neutral"  -- task describes a situation generic enough that users with very different
               traits would plausibly send it; exposure would have to come from the user's
               own reply, not from the task's wording.
- "fishing"  -- the task echoes or paraphrases a specific trait the sender is expected to
               have; reading the task alone lets you predict the sender's stance. Includes:
               (a) naming the trait ("how much detail is needed" -> privacy),
               (b) paraphrasing the trait ("fallback plan that doesn't get messy" -> low
               disruption), (c) soliciting self-description ("write a statement that sounds
               like me"), (d) encoding the trait's content (a note-taking template that
               pre-lists dates/numbers/times), (e) adjectives matching the expected stance
               ("make it sound calm and sensible" when target is "steady and practical").

Output JSON:
{
  "inferred_trait":  "<one-sentence guess at the trait / stance the task telegraphs, from the task alone>",
  "verdict":         "neutral" | "fishing",
  "why":             "<short reason>"
}
No commentary outside the JSON.
\end{Verbatim}
\end{promptbox}

\subsection{Anti-fishing editor prompt}
\label{app:prompt_antifishing_editor}
This is the system prompt for the editor that rewrites tasks flagged \emph{fishing} by the critic.
\begin{promptbox}
\begin{Verbatim}[fontsize=\scriptsize,breaklines=true,breakanywhere=true,breakautoindent=false,breakindent=0pt,breaksymbolleft={},breaksymbolsepleft=0pt]
You are a task editor. Rewrite a task that was flagged as FISHING so it becomes NEUTRAL
while still creating a situation where the target memory entry will naturally emerge in
the user's reply.

Preserve:
- The general scene / activity domain (unless the scene ITSELF is what encodes the trait).
- The situational trigger that would invite the trait to surface.

Remove:
- Any vocabulary that names, paraphrases, or presupposes the target trait's stance.
- Any phrasing that solicits self-description ("sound like me", "fit my style", etc.).
- Any list / template / structure that pre-encodes the target's content.

The rewritten task must describe a SITUATION; the user's reply (informed by THIS user's
profile) does the exposing. A reader seeing only the task text must not be able to guess
the target trait.

Output JSON:
{
  "task":      "<rewritten first-message text; 1-3 sentences>",
  "rationale": "<one sentence: why this neutral situation still elicits the target in the user's reply>"
}
No commentary outside the JSON.
\end{Verbatim}
\end{promptbox}

\subsection{User simulator prompt}
\label{app:prompt_simulator}
This is the system prompt for the persona-conditioned user simulator in \texttt{simulation.py}. The phrase ``specific real person'' in the prompt is used to discourage generic simulator responses; the benchmark users are synthetic profiles, as described in Appendix~\ref{app:benchmark_card}.
\begin{promptbox}
\begin{Verbatim}[fontsize=\scriptsize,breaklines=true,breakanywhere=true,breakautoindent=false,breakindent=0pt,breaksymbolleft={},breaksymbolsepleft=0pt]
You are roleplaying as a specific real person talking to an AI assistant.
Your ONLY job is to respond exactly as that person would -- not as a helpful, polite ideal user.

==== STEP 1: BECOME THIS PERSON ====
Read their base_profile and memory_bank carefully. Before writing anything, ask yourself:
- How does this person actually talk? (formal/casual, verbose/terse, which language mix)
- What are their emotional tendencies from episodic_memory and self_model?
- What would genuinely frustrate, excite, or confuse them given their skill_memory?
- What speech habits, filler words, or cultural patterns fit their background?

ANTI-ROBOT RULES -- violations make the simulation worthless:
- Do NOT start with "Thank you" or "Thanks" every turn. Real people don't do this.
- Do NOT use identical sentence structures across turns.
- Do NOT write polished, complete sentences if this person wouldn't. Fragments, hedges, and
  run-ons are fine if they fit the profile.
- Do NOT be uniformly positive. Show impatience, confusion, mild frustration, or genuine
  delight when the situation calls for it.
- Let the person's background bleed into the text: vocabulary level, cultural references,
  language switching, indirect communication style -- whatever fits who they are.

PROACTIVE AUTOBIOGRAPHICAL RECALL:
If the agent's current scenario closely parallels a specific past event in your
'episodic_memory' (a storm you lived through, a dispute you won, a scare you had, a
workplace conflict, a financial hit, a health issue), volunteer it briefly when it
would naturally come up in real speech: "I had something like this back when <brief
paraphrase>..." or "last time this happened, <what actually happened / what you
learned>". Don't wait for the agent to ask -- real people tell relevant stories without
being prompted.

Keep it in character (terse if this person is terse) and never dump the whole entry
verbatim; speak it the way this person would. Only recall when it genuinely parallels
the current situation -- don't shoehorn.

==== STEP 2: CLASSIFY THE REACTION ====
After deciding how the person feels, pick the feedback_type that best labels it:

- correction   -> agent stated a verifiable fact that contradicts your knowledge_memory
- add_context  -> agent's direction is off because it doesn't know something about you yet;
                 you supply the missing constraint (NOT for preference violations already known)
- too_deep     -> content exceeds your skill_memory / knowledge_memory
- too_shallow  -> content is far below your level, obvious or patronizing
- followup     -> direction is right, you want more depth or a sub-question in the same scope
- expose_gap   -> the response surfaced something in the material you don't understand
- satisfied    -> you RECEIVED the concrete deliverable (rewritten text, the list, the plan,
                 the specific answer), it meets your standards, AND nothing follow-up-worthy
                 remains. A promise to help, a clarifying question from the agent, a generic
                 outline, or an offer to produce something "if you paste X" is NOT satisfied.

==== STEP 3: WHEN THE AGENT IS WAITING ON YOU ====
If your opening message referenced something the agent needs to see in order to help (a draft,
a line, a letter, a note, an error message, a photo, a file, a price, a date, a document),
and the agent now asks you to share / paste / describe it, you MUST improvise a short
persona-consistent version of that content in your next reply -- fabricate it on the spot.

Do NOT end the turn saying "send it over", "go ahead", "ok", "yeah that works" while the
agent is still blocked on your input. 'satisfied' + 'is_done=true' is reserved for when
YOU'VE received what YOU asked for, not when the agent is waiting on material from you.
In this situation, use 'add_context' (you're providing the missing content) and set
'is_done=false'.

==== STEP 4: HARD BAR FOR SATISFIED -- READ BEFORE PICKING IT ====
'satisfied' is the narrowest feedback type -- treat it as a LAST RESORT. Before picking it,
verify ALL of these:
  (1) The agent has DELIVERED the concrete thing you asked for -- the rewritten text, the
      actual list, the specific plan, the direct answer. Not a promise. Not an offer
      conditional on "if you paste X". Not a meta-outline of what they will do.
  (2) The delivery matches your persona's standards (correctness, detail level, tone).
  (3) No natural follow-up, caveat, or refinement is pulled from your profile.

If ANY condition fails, pick one of the OTHER six types instead (all take 'is_done=false'):
  * agent only promised / asked for input     -> add_context  (supply what they need)
  * delivery missed a constraint from your profile -> add_context
  * delivery is shallow / patronizing          -> too_shallow
  * delivery is over your head                  -> too_deep
  * right direction but you want more          -> followup
  * stated fact contradicts your knowledge     -> correction
  * delivery surfaced your own knowledge gap   -> expose_gap

ANTI-PREMATURE-SATISFACTION: First-turn satisfaction is rare. Real help conversations average
2-5 turns. If you're tempted to say satisfied on turn 1, it's almost always because the agent
promised help without delivering, or gave a generic overview. Push back with add_context or
followup and set 'is_done=false'.

==== OUTPUT ====
Return a JSON object -- no markdown, no extra keys:
{
  "feedback_type": "<one of the seven values above>",
  "text": "<your response as this person>"
}
\end{Verbatim}
\end{promptbox}

\subsection{Reconstruction slot-fill prompt}
\label{app:prompt_slotfill}
This is the system prompt for the slot-fill step that produces a per-dimension prediction from the memory store.
\begin{promptbox}
\begin{Verbatim}[fontsize=\scriptsize,breaklines=true,breakanywhere=true,breakautoindent=false,breakindent=0pt,breaksymbolleft={},breaksymbolsepleft=0pt]
An AI assistant had a series of conversations with a specific user and stored the following memory notes.
Each note preserves the full structure recorded by the memory system.

Memory notes (JSON array):
{memories_json}

Based solely on the above, fill in this dimension about the user.

Memory category: {category}
Category means:  {category_guide}

Dimension: {dimension}

Answer STRICTLY on the dimension within the category. Do NOT drift into another category:
- If the category is skill_memory, describe what they can DO, not what they prefer or believe.
- If the category is assistance_preference, describe delivery style, not their skills or knowledge.
- Same discipline for the other categories -- the "NOT" clauses above are binding.

Output format requirements:
- "label" <=15 words, direct and declarative. No "The user..." framing, no hedging like "seems to".
- "reason" is one short sentence citing the specific memory evidence you relied on (for audit);
  if no evidence supports a concrete answer, reason explains that and label is "unknown".

Label guidance (applies to the "label" field):
  Good:  "struggles with algebra, relies on calculators"
  Bad:   "The user seems to have some difficulty with algebra and appears to rely on calculators when solving problems."
  Wrong category drift: "prefers step-by-step explanations" (<- that is assistance_preference, not skill)

Output JSON only -- no markdown fences, no commentary:
{{"label": "<terse value>", "reason": "<one-sentence evidence citation>"}}
\end{Verbatim}
\end{promptbox}

\subsection{Reconstruction judge prompt}
\label{app:prompt_judge}
This is the system prompt for the recovery judge that scores predictions against the hidden ground truth on the five-level rubric.
\begin{promptbox}
\begin{Verbatim}[fontsize=\scriptsize,breaklines=true,breakanywhere=true,breakautoindent=false,breakindent=0pt,breaksymbolleft={},breaksymbolsepleft=0pt]
Compare two short descriptions of the same person on the same dimension.

Dimension: {dimension}
What this dimension asks about: {explanation}

Ground truth: {ground_truth}
Predicted:    {predicted}

Score on a 5-level scale (0.0 - 1.0):
  1.00 -- exact: captures the same meaning completely, no contradictions or omissions
  0.75 -- close: same core meaning, minor aspects missing or loosely phrased
  0.50 -- partial: some overlap with ground truth, but key elements missing or slightly off
  0.25 -- weak:    marginal overlap; mostly misses the point
  0.00 -- wrong:  contradicts ground truth, entirely unrelated, or predicted is "unknown"

Output JSON only: {{"score": <0.0|0.25|0.5|0.75|1.0>, "reason": "<one sentence>"}}
\end{Verbatim}
\end{promptbox}

\subsection{Failure-attribution Stage~0 (task-design oracle)}
\label{app:prompt_oracle}
This is the task-design oracle. Cached per \((user, dimension)\) and shared across runs.
\begin{promptbox}
\begin{Verbatim}[fontsize=\scriptsize,breaklines=true,breakanywhere=true,breakautoindent=false,breakindent=0pt,breaksymbolleft={},breaksymbolsepleft=0pt]
You are deciding whether a single task can ever naturally invite a user to
reveal information about a target dimension. Assume a perfectly probing
assistant AND a perfectly cooperative user simulator. You are NOT looking
at any actual conversation -- judge purely from the task statement and
the dimension being asked about.

Target dimension: {dimension}
Ground truth (short):   {gt_short}
Ground truth (full):    {gt_explanation}

Task:
{task_text}

Could this task plausibly lead -- through reasonable follow-up questions
and natural elaboration -- to a moment where the user reveals their value
on the target dimension?

Output JSON only (no markdown):
{{"can_invite": true|false, "reason": "<one sentence>"}}\
\end{Verbatim}
\end{promptbox}

\subsection{Failure-attribution Stage~1 (disclosure check)}
\label{app:prompt_disclosure}
This is the disclosure judge. Run on dims with score \(<0.75\) that pass the oracle (\code{can\_invite}=true).
\begin{promptbox}
\begin{Verbatim}[fontsize=\scriptsize,breaklines=true,breakanywhere=true,breakautoindent=false,breakindent=0pt,breaksymbolleft={},breaksymbolsepleft=0pt]
You are checking whether one episode of dialogue contains enough information
about a target dimension that a competent memory system could later infer
the user's value on this dimension.

Target dimension: {dimension}
Ground truth (short):   {gt_short}
Ground truth (full):    {gt_explanation}

The episode is a multi-turn assistance dialogue between an Agent and a User
simulator playing a fixed persona. Read it carefully and decide:

  Did the User simulator (or the Agent quoting the user) reveal enough
  about the target dimension that a competent memory system, after
  reading the dialogue, could plausibly arrive at the user's value on
  this dimension?

Important guidance:
- You are NOT asking whether the literal ground-truth phrasing was uttered.
- Indirect signals count: behavioural hints, contextual mentions, clear
  preferences expressed in passing, implications of what the user said.
- If a reasonable downstream reader would, given the dialogue, conclude
  something close to the ground truth (even if loosely worded), answer YES.
- Only answer NO if the dialogue offers essentially no signal a reader
  could anchor on.

Episode transcript:
================================
TASK: {task_text}

{transcript}
================================

Output JSON only (no markdown):
{{"disclosed": true|false, "evidence": "<short quote or 'none'>", "reason": "<one sentence>"}}\
\end{Verbatim}
\end{promptbox}

\subsection{Failure-attribution Stage~2 (subclass)}
\label{app:prompt_subclass}
This is the agent-vs.-simulator subclassifier. Run on dims that pass Stage~0 but fail Stage~1 (\code{disclosed}=false).
\begin{promptbox}
\begin{Verbatim}[fontsize=\scriptsize,breaklines=true,breakanywhere=true,breakautoindent=false,breakindent=0pt,breaksymbolleft={},breaksymbolsepleft=0pt]
The dialogue below failed to disclose the target dimension's ground truth.
We have already verified that the task COULD invite this dimension under
perfect conditions, so the failure must come from how the conversation
actually unfolded. Decide which side is responsible.

Target dimension: {dimension}
Ground truth (short):   {gt_short}
Ground truth (full):    {gt_explanation}

Pick exactly one:
  A. agent_elicitation_failure
     The agent never went anywhere near this dimension. The conversation
     stayed on unrelated territory; the agent did not roughly steer toward
     the topic this dim is about.

  B. simulator_too_strict
     The agent did roughly steer the conversation toward this dim's topic
     area (does not need to be a literal pinpoint question), but the user
     simulator did not actually disclose enough -- gave shallow / non-
     committal answers, changed direction, declared satisfied early, or
     simply did not elaborate when the topic was on the table.

A rough rule of thumb: if reading the transcript you feel "the topic came
up, the user just didn't say much about it" -> B. If you feel "the agent
never even brought this kind of thing up" -> A.

Episode transcript:
================================
TASK: {task_text}

{transcript}
================================

Output JSON only (no markdown):
{{"category": "A|B", "reason": "<one sentence>", "evidence": "<short quote>"}}\
\end{Verbatim}
\end{promptbox}

%% file: appendix/scoring_details.tex
\section{Scoring and Attribution Protocol Details}
\label{app:scoring_details}

\paragraph{Recovery score $B$.} For each $(\text{user}, \text{dimension})$, the slot-filler
predicts \code{\{label, reason\}} from the memory evidence available under
the chosen access mode; the recovery judge then scores the predicted label
against the ground-truth \code{short} (with the dimension \code{explanation}
visible to the judge) on the five-level rubric in
Table~\ref{tab:reconstruction_rubric}. For each user, the headline $B$ score is
category-balanced: the $31$ dimension-level scores are first averaged within the
five memory categories, and the five category means are then averaged with equal
weight. Tables~\ref{tab:main} and~\ref{tab:main_percat} report the per-system
mean and standard deviation across $50$ users.

\paragraph{Access modes.} In \code{dump\_all}, the slot-filler receives the full
memory store via \code{agent.get\_all\_memories()}, JSON-serialized as a list
of memory items preserving each system's native fields (note id, content,
metadata, links / collection name where applicable). In \code{retrieve}, the
slot-filler receives only \code{agent.search(query, k=5)} for an axis-only
query of the form ``What best describes the user's \code{<dimension>}?''
followed by a fixed category guide (skill / knowledge / episodic / self /
preference). The query intentionally contains no user-specific content; this is
what stresses the read interface and leaves the query-construction policy out of scope.

\paragraph{Sampling and call caching.}
The slot-filler, recovery judge, task-design oracle, disclosure judge, and
subclass judge all run at \code{temperature=0.0} to minimize sampling
variance. Outputs are typically stable across reruns but not bit-deterministic,
because API-side numerical precision and inference infrastructure introduce
small non-deterministic variation even at \code{temperature=0.0}; we
therefore treat the judges as low-noise estimators whose outputs may vary
slightly, report aggregate scores over $50$ users $\times$ $31$ dims, and
release every judge input, output, and rationale so individual decisions can
be inspected. The Stage-0 oracle is keyed on $(\text{user\_id}, \text{dimension})$ and
cached across runs, so a given $(\text{user}, \text{dimension})$ has a single Stage-0
verdict reused across all five systems and both access modes. Thus,
when a case reaches Stage~0, differences in the oracle verdict are not due
to stochastic re-judging; the row counts still depend on which cases enter
the pipeline after recovery thresholding. The disclosure and subclass judges are not cached
because they read the per-run transcript, which differs per system.

\paragraph{Interpreting attribution counts across access modes.}
Attribution counts are not directly comparable across \code{dump\_all} and \code{retrieve} as raw frequencies of task or simulator behavior. The attribution pipeline first thresholds recovery: cases with score $\geq 0.75$ are labeled \code{ok} and bypass later stages. As a result, when retrieval lowers a case below threshold, it can newly enter the Stage~0--2 attribution pipeline and receive labels such as \code{agent\_elicitation\_failure} or \code{simulator\_too\_strict}. Thus, higher \code{agtE} or \code{simS} counts in retrieve rows should not be read as evidence that the underlying task or simulator changed under retrieval. They indicate labels assigned among the subset of cases that became unrecovered under that access mode.

\paragraph{Threshold rationale.} The $0.75$ threshold corresponds to
``mostly correct reconstruction with minor missing detail or phrasing
mismatch'' (Table~\ref{tab:reconstruction_rubric}); below this, the prediction
either misses the central target value, drifts to a neighboring abstraction,
or returns \code{unknown}. The attribution pipeline short-circuits any dim
with $B \geq 0.75$ to \code{ok} so that disclosed dims with confidently correct
predictions do not consume Stage~1/2 judge calls. This also produces the
asymmetry between \code{dump\_all} and \code{retrieve} attribution
populations discussed in the Results and Analysis section.

\input{tables/tab_reconstruction_rubric}

\input{tables/tab_attrib_rules}

%% file: tables/tab_reconstruction_rubric.tex
\begin{table}[t]
	\caption{Reconstruction scoring rubric used by the recovery judge.}
	\label{tab:reconstruction_rubric}
	\centering
	\small
	\setlength{\tabcolsep}{4pt}
	\begin{adjustbox}{max width=\linewidth}
		\begin{tabular}{@{}cp{0.72\linewidth}@{}}
			\toprule
			Score & Interpretation \\
			\midrule
			0.00 & No recoverable match to the target dimension. \\
			0.25 & Weak topical overlap, but the central target value is missing or wrong. \\
			0.50 & Partial match with important omissions, abstraction drift, or slot-boundary errors. \\
			0.75 & Mostly correct reconstruction with minor missing detail or phrasing mismatch. \\
			1.00 & Correct reconstruction of the target value at the intended abstraction level. \\
			\bottomrule
		\end{tabular}
	\end{adjustbox}
\end{table}

%% file: tables/tab_attrib_rules.tex
\begin{table}[t]
	\caption{Failure-attribution decision rules.}
	\label{tab:attrib_rules}
	\centering
	\small
	\setlength{\tabcolsep}{4pt}
	\begin{adjustbox}{max width=\linewidth}
		\begin{tabular}{@{}p{0.34\linewidth}p{0.56\linewidth}@{}}
			\toprule
			condition & attribution label \\
			\midrule
			Task cannot plausibly invite the target dimension & \code{task\_design\_failure} \\
			Task can invite, but the dialogue never approaches the target topic & \code{agent\_elicitation\_failure} \\
			Topic appears, but the simulator does not disclose enough signal & \code{simulator\_too\_strict} \\
			Signal is disclosed, but recovery score is below $0.75$ & \code{memory\_failure} \\
			Signal is disclosed and recovery score is at least $0.75$ & \code{ok} \\
			\bottomrule
		\end{tabular}
	\end{adjustbox}
\end{table}

%% file: appendix/human_validation.tex
\section{Human Validation of Pipeline Judgments}
\label{app:human_validation}

The headline metrics in Table~\ref{tab:main} and the attribution counts in Table~\ref{tab:attrib} rely on automated simulator and attribution judgments. We therefore run a small human audit over stratified samples of both signals, with results in Tables~\ref{tab:human_satisfied} and~\ref{tab:human_attribution}. The simulator's \code{satisfied} signal is judged reasonable in $48{/}50$ cases ($96\%$), supporting the use of \code{end\_reason} as the basis of $A$. Failure-attribution labels agree with human re-labels in $90{/}120$ cases ($75.0\%$). The remaining disagreements are limited and mostly occur between neighboring failure categories: the overall human-label distribution remains similar across the four failure modes (Memory $29$, Task $25$, Agent $25$, User $32$ human labels, against $30$ automated each), and the \emph{Memory} mode is nearly unchanged in aggregate ($30$ automated samples vs.\ $29$ human labels). Together, these results suggest that the automated judgments are broadly reliable for our analysis, while preserving the expected caveat that some boundary cases remain ambiguous.

\input{tables/tab_human_satisfied}

\input{tables/tab_human_attribution}

\paragraph{Binary disclosure reanalysis.}
For the central distinction between absent evidence and
post-disclosure failure, we collapse the same human annotations into a
binary disclosed-versus-not-disclosed decision
(Table~\ref{tab:human_attribution}, panel~(b)).
Automated \code{memory\_failure} maps to \emph{disclosed};
automated \code{agent\_elicitation\_failure} and
\code{simulator\_too\_strict} map to \emph{not disclosed}.
Human \emph{Memory} and \emph{Recovered} labels map to
\emph{disclosed}; human \emph{Agent} and \emph{User} labels map to
\emph{not disclosed}. Task-design and indeterminate cases are
excluded.

Automated and human decisions agree in $71/86$ applicable cases
($82.6\%$, Cohen's $\kappa{=}0.61$).
Among the $28$ applicable cases in the automated
\code{memory\_failure} stratum, $22$ are human-confirmed as disclosed;
equivalently, $22$ of the full $30$ sampled
\code{memory\_failure} cases contain human-confirmed disclosure.
This is a reanalysis of the existing annotations rather than an
additional annotation round.

\paragraph{Target observability.}
We define a target as observable when it is either recovered or judged
disclosed but unrecovered. Under \code{dump\_all}, automated
observability is $94.4\%$ for \code{longctx\_full}, $93.9\%$ for
\code{mem0}, and $93.5\%$ for \code{amem}.

Applying the human-confirmed precision of the automated disclosed
decision ($22/28$) to the disclosed-but-unrecovered counts yields
audit-adjusted estimates of $87.7\%$, $87.0\%$, and $86.6\%$,
respectively.

\paragraph{Attribution-judge sensitivity.}
We additionally rerun attribution with two alternative judges while
holding the underlying cases fixed.
Table~\ref{tab:attribution_judge} shows that
$91.6\%$--$96.6\%$ of attributed failures fall on the combined
agent-or-memory side, compared with $3.4\%$--$8.4\%$ assigned to task
design or simulator strictness. The main variation is in how judges
divide system-side cases between \code{memory\_failure} and
\code{agent\_elicitation\_failure}.

\input{tables/tab_attribution_judge}

Taken together, these analyses support the disclosed-versus-undisclosed
distinction and the aggregate agent-or-memory-side diagnosis used in
the main text.

%% file: tables/tab_human_satisfied.tex
\begin{table}[t]
\centering
\small
\caption{Human validation of the simulator's \code{satisfied} signal across $50$ stratified episodes.}
\label{tab:human_satisfied}
\setlength{\tabcolsep}{6pt}
\begin{tabular}{@{}lcccc@{}}
\toprule
End state & Reasonable & Borderline & Unreasonable & Total \\
\midrule
\code{satisfied} & \textbf{48} (96\%) & 1 (2\%) & 1 (2\%) & 50 \\
\bottomrule
\end{tabular}
\end{table}

%% file: tables/tab_human_attribution.tex
\begin{table}[t]
\centering
\small
\caption{Human validation of failure-attribution labels (rows: original automated label; columns: human re-label; $30$ samples per row, diagonal in bold). Categories follow Table~\ref{tab:attrib}.}
\label{tab:human_attribution}
\setlength{\tabcolsep}{6pt}
\begin{tabular}{@{}l cccc cc c@{}}
\toprule
 & \multicolumn{6}{c}{Human re-label} & \\
\cmidrule(lr){2-7}
Original label
 & Memory & Task & Agent & User & Recovered & Other & Total \\
\midrule
Memory & \textbf{20} & 1 & 1 & 5 & 2 & 1 & 30 \\
Task   & 3 & \textbf{23} & 0 & 2 & 2 & 0 & 30 \\
Agent  & 1 & 1 & \textbf{24} & 2 & 2 & 0 & 30 \\
User   & 5 & 0 & 0 & \textbf{23} & 1 & 1 & 30 \\
\midrule
Column total & 29 & 25 & 25 & 32 & 7 & 2 & 120 \\
\bottomrule
\end{tabular}

\vspace{6pt}

\textbf{(b) Binary disclosure reanalysis}\\[-2pt]

\setlength{\tabcolsep}{9pt}
\begin{tabular}{@{}lrrr@{}}
\toprule
& \multicolumn{2}{c}{Human decision} & \\
\cmidrule(lr){2-3}
Automated decision
& Disclosed
& Not disclosed
& Total
\\
\midrule
Disclosed
& \textbf{22}
& 6
& 28
\\
Not disclosed
& 9
& \textbf{49}
& 58
\\
\midrule
Total
& 31
& 55
& 86
\\
\bottomrule
\end{tabular}

\end{table}

%% file: tables/tab_attribution_judge.tex
\begin{table}[H]
\centering
\small
\caption{
Failure-attribution distribution under alternative judges.
System-side is the sum of \emph{Memory} and \emph{Elicitation}.
}
\label{tab:attribution_judge}

\setlength{\tabcolsep}{5pt}
\begin{tabular}{@{}lrrrrr@{}}
\toprule
Judge & Memory & Elicitation & Simulator & Task design & System-side \\
\midrule
\code{gpt-5.4-mini} & 82.6\% & 9.0\% & 2.3\% & 6.1\% & \textbf{91.6\%} \\
\code{gpt-5.4} & 70.8\% & 25.8\% & 1.0\% & 2.4\% & \textbf{96.6\%} \\
\code{gpt-5.5} & 68.7\% & 25.1\% & 6.2\% & 0.0\% & \textbf{93.8\%} \\
\bottomrule
\end{tabular}
\end{table}

%% file: appendix/additional_results.tex
\section{Additional Results}
\label{app:additional_results}
\paragraph{Per-category attribution counts (\code{dump\_all}).} Table~\ref{tab:attrib_percat}
breaks Table~\ref{tab:attrib} down by the five recovery categories. The dominant
non-\code{ok} label is \code{memory\_failure} for every system and category,
confirming that the headline gap lies on the write/recovery side, downstream of
disclosure. \cat{Episodic Memory} contributes a large share of
\code{agtE} cases and many \code{taskD} cases, while \cat{Self Model}
also contributes many \code{taskD} cases. This is consistent with the
difficulty of inviting time-anchored autobiographical events and abstract
self-views through ordinary assistance dialogue.
\input{tables/tab_attrib_percat}
\paragraph{Per-user dump\,$-$\,retrieve gap.} For the three context-fit systems
(\code{amem}, \code{longctx\_full}, \code{mem0}) the dump\,$-$\,retrieve $B$ gap
is positive and reasonably tight; here $p_{50}$ denotes the standard median over
$50$ users (the footprint quantiles in Table~\ref{tab:footprint_dist} instead
follow the script's lower-order-statistic convention): \code{amem} $+0.071$ mean
($p_{50}{=}+0.075$, range $[-0.026, +0.174]$), \code{longctx\_full}
$+0.121$ mean ($p_{50}{=}+0.129$, range $[-0.001, +0.256]$), \code{mem0}
$+0.140$ mean ($p_{50}{=}+0.149$, range $[-0.044, +0.319]$). \code{memt} is the
expected outlier: $-0.335$ mean ($p_{50}{=}-0.456$, range $[-0.574, +0.234]$),
because retrieve actually outperforms its overflowed \code{dump\_all} probe on
most users. The handful of negative entries for the context-fit systems (a few
users where retrieve happened to surface a slightly better top-$5$ than the
full store) confirms that the gap is not a uniform artifact of the slot-fill
prompt.

We isolate scoring-layer dependence by holding benchmark instances,
dialogues, memory artifacts, retrieval outputs, and reconstruction
predictions fixed while varying only the recovery judge or the wording
of the scoring prompt.

\subsection{Robustness of the Main Results}
\label{app:robustness_checks}

\paragraph{Recovery-judge swaps.}
Table~\ref{tab:recovery_judge_scores} reports the full per-system
scores under two same-family and two cross-family judges.
All four alternative judges preserve \code{longctx\_full} as the
\code{dump\_all} winner and \code{amem} as the \code{retrieve}
winner. Absolute calibration shifts across judges, with the
cross-family judges producing a broadly lower scale, while all
ordering changes are confined to original near-ties below $0.03$.
Every larger-gap comparison remains unchanged
($22/22$ system comparisons and $69/69$ category comparisons), and
all judges preserve the $8/8$ hardest/easiest category positions.

\input{tables/tab_recovery_judge_scores}

\paragraph{Prompt paraphrases.}
Table~\ref{tab:prompt_robustness} reports two rubric-preserving
paraphrases of the recovery-scoring prompt. Both preserve the complete
ordering in each access mode, both access-mode winners, and every
hardest/easiest category position. Variant~A leaves aggregate
calibration nearly unchanged, whereas Variant~B produces a broadly
uniform downward shift. Thus, prompt wording affects absolute
calibration and some near-ties without changing the headline
comparisons.

\input{tables/tab_prompt_robustness}

\subsection{Downstream PersonaGym Validation}
\label{app:personagym}
\paragraph{Protocol.}
We use PersonaGym \citep{personagym} as a complementary downstream
evaluation of whether recoverable memory supports subsequent
persona-dependent behavior. Ground-truth persona information is used
only to construct questions and grading targets and is hidden from the
answering model.
For each question, the answering model receives the system's native
top-$5$ retrieved memories as its only user-specific context;
\code{nomem} receives no user-specific memory. We retain PersonaGym's
task construction, score exemplars, rubric-based grading, and
two-evaluator ensemble. Across $50$ users, five task types, ten
questions per task, and four evaluated settings, the study contains
$10{,}000$ answers.
\paragraph{Evaluation coverage.}
We completed and verified
\code{nomem}, \code{longctx\_full}, \code{amem}, and \code{mem0}.
The corresponding \code{memt} run was not completed.
The protocol is also applicable to \code{memt}; its omission reflects
evaluation coverage rather than a methodological incompatibility.
We therefore report a four-setting downstream validation and make no
PersonaGym claim for \code{memt}.
\paragraph{Results.}
Table~\ref{tab:personagym_main} reports the results.
Memory-3 is the descriptive mean across Linguistic Habits, Persona
Consistency, and Action Justification. Every evaluated
memory-equipped setting exceeds \code{nomem} on Memory-3.
Retrieve-mode recovery, PersonaScore, and Memory-3 exhibit the same
numerical ordering:
\[
\code{amem} > \code{longctx\_full} > \code{mem0} > \code{nomem}.
\]
Because the differences among the three memory-equipped settings are
small, particularly for PersonaScore, we interpret this alignment as
complementary evidence for the behavioral relevance of memory
recovery, rather than as a statistically established pairwise
ranking.

%% file: tables/tab_attrib_percat.tex
\begin{table}[t]
	\caption{
            Failure-attribution counts by user-state category under dump\_all over 50 users.
            }
	\label{tab:attrib_percat}
	\centering
	\small
	\setlength{\tabcolsep}{4pt}
	\begin{adjustbox}{max width=\linewidth}
		\begin{tabular}{@{}llrrrrr@{}}
			\toprule
			System & Category & Recovered\,$\uparrow$ & Memory\,$\downarrow$ & Task\,$\downarrow$ & Agent\,$\downarrow$ & User\,$\downarrow$ \\
			\midrule
			\code{nomem}   & sk   &   0 & 347 &   0 &  2 & 1 \\
			               & kn   &   0 & 342 &   2 &  5 & 1 \\
			               & ep   &   0 & 283 &  28 & 37 & 2 \\
			               & self &   0 & 194 &  38 & 17 & 1 \\
			               & pref &   0 & 244 &   5 &  1 & 0 \\
			\midrule
			\code{amem}    & sk   & 229 & 120 &   0 &  1 & 0 \\
			               & kn   & 237 & 109 &   0 &  4 & 0 \\
			               & ep   & 104 & 186 &  23 & 33 & 4 \\
			               & self & 148 &  68 &  21 & 13 & 0 \\
			               & pref & 232 &  17 &   1 &  0 & 0 \\
			\midrule
			\code{longctx\_full} & sk   & 241 & 105 &   0 &  3 & 1 \\
			               & kn   & 241 & 106 &   0 &  3 & 0 \\
			               & ep   & 110 & 189 &  22 & 28 & 1 \\
			               & self & 164 &  58 &  20 &  7 & 1 \\
			               & pref & 226 &  23 &   1 &  0 & 0 \\
			\midrule
			\code{mem0}    & sk   & 245 & 104 &   0 &  1 & 0 \\
			               & kn   & 226 & 119 &   0 &  3 & 2 \\
			               & ep   & 108 & 184 &  23 & 34 & 1 \\
			               & self & 159 &  60 &  21 &  8 & 2 \\
			               & pref & 223 &  27 &   0 &  0 & 0 \\
			\midrule
			\code{memt}    & sk   &  45 & 304 &   0 &  0 & 1 \\
			               & kn   &  47 & 297 &   2 &  4 & 0 \\
			               & ep   &  22 & 260 &  28 & 36 & 4 \\
			               & self &  32 & 166 &  34 & 18 & 0 \\
			               & pref &  51 & 192 &   5 &  2 & 0 \\
			\bottomrule
		\end{tabular}
	\end{adjustbox}
\end{table}

%% file: tables/tab_recovery_judge_scores.tex
\begin{table}[!t]
\centering
\footnotesize
\caption{Category-balanced recovery $B$ under alternative recovery judges (mean $\pm$ standard deviation over $50$ users). Benchmark instances, dialogues, memory artifacts, retrieval outputs, and reconstruction predictions are held fixed; only the judge changes. Bold indicates the best system within each access mode and judge.}
\label{tab:recovery_judge_scores}
\setlength{\tabcolsep}{3.5pt}
\begin{adjustbox}{max width=\linewidth}
\begin{tabular}{@{}llccccc@{}}
\toprule
Mode & System & \shortstack{Released\\(gpt5.4mini)} & gpt5.4 & gpt5.5 & \shortstack{Claude\\Sonnet 5} & \shortstack{Claude\\Opus 4.8} \\
\midrule
\code{dump\_all} & \code{amem} & $0.611 \pm 0.062$ & $0.620 \pm 0.063$ & $0.626 \pm 0.075$ & $0.568 \pm 0.090$ & $0.575 \pm 0.078$ \\
 & \code{longctx\_full} & $\mathbf{0.624 \pm 0.067}$ & $\mathbf{0.629 \pm 0.069}$ & $\mathbf{0.636 \pm 0.076}$ & $\mathbf{0.582 \pm 0.094}$ & $\mathbf{0.592 \pm 0.083}$ \\
 & \code{mem0} & $0.613 \pm 0.060$ & $0.617 \pm 0.062$ & $0.622 \pm 0.066$ & $0.569 \pm 0.086$ & $0.573 \pm 0.076$ \\
 & \code{memt}$^{\dagger}$ & $0.130 \pm 0.251$ & $0.132 \pm 0.254$ & $0.136 \pm 0.261$ & $0.123 \pm 0.240$ & $0.124 \pm 0.241$ \\
\midrule
\code{retrieve} & \code{amem} & $\mathbf{0.540 \pm 0.062}$ & $\mathbf{0.541 \pm 0.066}$ & $\mathbf{0.544 \pm 0.065}$ & $\mathbf{0.473 \pm 0.078}$ & $\mathbf{0.486 \pm 0.074}$ \\
 & \code{longctx\_full} & $0.503 \pm 0.075$ & $0.510 \pm 0.074$ & $0.510 \pm 0.072$ & $0.441 \pm 0.087$ & $0.453 \pm 0.078$ \\
 & \code{mem0} & $0.473 \pm 0.079$ & $0.474 \pm 0.079$ & $0.475 \pm 0.077$ & $0.410 \pm 0.081$ & $0.419 \pm 0.077$ \\
 & \code{memt} & $0.465 \pm 0.057$ & $0.470 \pm 0.060$ & $0.471 \pm 0.057$ & $0.414 \pm 0.066$ & $0.420 \pm 0.063$ \\
\bottomrule
\end{tabular}
\end{adjustbox}
\par\smallskip
\begin{minipage}{0.97\linewidth}
\footnotesize
$^{\dagger}$The \code{memt} \code{dump\_all} result remains a diagnostic full-store auditability outcome because the serialized store exceeds the recovery context; it is not directly comparable to the context-fit full-store rows. All four alternative judges preserve the winner in each access mode. Absolute scores vary across judges, whereas all comparison changes are confined to original gaps below $0.03$.
\end{minipage}
\end{table}

%% file: tables/tab_prompt_robustness.tex
\begin{table}[!t]
\centering
\small
\caption{Recovery-scoring robustness under two rubric-preserving prompt paraphrases.}
\label{tab:prompt_robustness}
\setlength{\tabcolsep}{8pt}
\begin{tabular}{@{}lcc@{}}
\toprule
Metric & Variant A & Variant B \\
\midrule
Per-target exact agreement & $85.7\%$ & $81.7\%$ \\
Within one rubric level ($|\Delta| \leq 0.25$) & $99.9\%$ & $99.7\%$ \\
Mean signed shift & $+0.0002$ & $-0.0277$ \\
Best system per access mode & $\mathbf{2/2}$ & $\mathbf{2/2}$ \\
System pairs preserved & $48/54$ & $47/54$ \\
Category pairs preserved & $79/80$ & $78/80$ \\
Hardest/easiest category positions & $\mathbf{8/8}$ & $\mathbf{8/8}$ \\
\bottomrule
\end{tabular}
\par\smallskip
\begin{minipage}{0.94\linewidth}
\small
Both variants preserve the complete ordering in each access mode. Variant~A leaves aggregate calibration nearly unchanged; Variant~B produces a broadly uniform downward shift. All changed comparisons are original near-ties below $0.03$.
\end{minipage}
\end{table}

%% file: appendix/recall.tex
\section{Per-system Retrieval and Budget Probes}
\label{app:recall}
\paragraph{Memory footprint distribution.}
Table~\ref{tab:footprint_dist} summarizes the per-user memory store size for
each memory-equipped system. Among the context-fit systems, \code{mem0} writes
the largest number of items ($p_{50}{=}190$) but the shortest items
($p_{50}{=}206$ chars), giving the smallest total store. \code{memt} writes both
the most items ($p_{50}{=}458$)
and the longest items ($p_{50}{=}1028$ chars); its $p_{90}$ user has
$596 \times 1288 \approx 768$\,K characters of raw memory content, before
metadata.
\input{tables/tab_footprint_dist}
\paragraph{Why \code{memt}'s \code{dump\_all} overflows.} The \code{dump\_all}
probe serializes the entire store as a JSON array and feeds it to the
slot-fill prompt. For \code{amem}, \code{longctx\_full}, and \code{mem0},
the resulting prompt fits the recovery agent's context window even on
$p_{90}$ users. For \code{memt}, the median user already produces a serialized
store of roughly $458 \times 1028 \approx 470$\,K content characters, plus
metadata for each typed collection (\code{turns / facts / experiences /
personas / summary}); this exceeds the slot-fill model's input limit on most
users. The result is the diagnostic outcome reported in
Tables~\ref{tab:main}--\ref{tab:attrib}: the slot-filler degrades to ``unknown''
on the affected users, and the per-system mean \code{dump\_all} $B$ collapses
to $0.130$ even though the underlying store does contain recoverable evidence
(visible to the same evaluator under \code{retrieve}, where only the top-$5$
items are passed in).
\paragraph{Retrieve interface budgets.}
All systems use $k{=}5$ for the headline retrieve numbers. \code{amem},
\code{longctx\_full}, and \code{mem0} return their top-$5$ from a single
embedding-kNN pool; \code{memt}'s \code{search} runs Mem-T-4B's ReAct loop
which can call any of \code{search\_facts / search\_experiences /
search\_turns / search\_summary / search\_personas} sequentially under a
$\code{max\_tool\_steps}{=}6$ budget, accumulates surfaced items across
steps, and returns the top-$5$ after de-duplication. The released artifact
includes the actual top-$5$ items returned for every $(\text{user}, \text{dimension})$ pair
so retrieval quality can be re-audited independently of the slot-fill step.
Top-$k$ sensitivity sweeps ($k \in \{1, 3, 5, 10\}$) are left as a planned
robustness probe; with the present $k{=}5$ headline, the retrieve-mode
ranking is \code{amem} $>$ \code{longctx\_full} $>$ \code{mem0} $>$ \code{memt},
matching Table~\ref{tab:main}.

%% file: tables/tab_footprint_dist.tex
\begin{table}[t]
	\caption{Per-user memory footprint distribution across $N{=}50$ users. ``$N$ items'' is the count returned by \code{agent.get\_all\_memories()}; ``chars/item'' is the mean character length of the \code{content} field per memory across that user's items. For even $N{=}50$, the $p_{50}$ and $p_{90}$ columns follow the lower-order-statistic quantile convention emitted by the released footprint aggregation script, which does not average the two middle values.}
	\label{tab:footprint_dist}
	\centering
	\small
	\setlength{\tabcolsep}{6pt}
	\begin{adjustbox}{max width=\linewidth}
		\begin{tabular}{@{}lrrrrrrrr@{}}
			\toprule
			& \multicolumn{4}{c}{$N$ items / user} & \multicolumn{4}{c}{chars / item} \\
			\cmidrule(lr){2-5}\cmidrule(lr){6-9}
			system & mean & $p_{50}$ & $p_{90}$ & max & mean & $p_{50}$ & $p_{90}$ & max \\
			\midrule
			\code{amem}          & 108.6 & 103 & 129 & 164 &  429.2 &  439.4 &  568.4 &  697.2 \\
			\code{longctx\_full} & 155.4 & 150 & 190 & 302 &  685.7 &  625.3 & 1001.6 & 1383.9 \\
			\code{mem0}          & 203.8 & 190 & 302 & 312 &  205.0 &  205.9 &  218.1 &  237.2 \\
			\code{memt}          & 471.0 & 458 & 596 & 643 & 1014.9 & 1027.6 & 1287.9 & 1510.5 \\
			\bottomrule
		\end{tabular}
	\end{adjustbox}
\end{table}

%% file: appendix/gallery.tex
\section{Failure case gallery}
\label{app:gallery}

Each box below packages one audited $(\text{user}, \text{dimension})$ case with the information
needed to trace the diagnosis back to released artifacts: target, task, diagnosis,
and, where relevant, abridged transcript or retrieved memories, per-system
predictions, and judge reasoning.

\textbf{Case selection.} We selected cases to cover three recurring mechanisms
observed in the audit: trace-to-target abstraction, write-side abstraction, and
query--content mismatch. Boundary cases are included when they expose
limitations of the scoring or retrieval analysis without changing the aggregate
conclusions. Specifically, Cases~1, 2, and 4 instantiate the three main
mechanisms; Cases~3, 5, and 6 are boundary conditions (a slot-label polarity
artifact, a localized retrieval advantage of typed memory collections, and rare
turn-budget task failures), included to keep the audit honest, without
expanding the paper's main claims.

% ─────────────────────────────────────────────────────────────────────────
\begin{failbox}{Case 1: trace-to-target abstraction in episodic memory\quad\textnormal{\code{user\_010 / moment\_of\_clarity\_event}}}
\small
\textbf{Role in audit.} Main mechanism: trace-to-target abstraction.\quad
\textbf{Pattern.} Universal-low: every non-\code{memt} system scores below $0.5$
in \code{dump\_all}.\quad
\textbf{Category.} \cat{Episodic Memory}.

\textbf{Ground truth.} ``Realized questions help more than pretending.''
\emph{Explanation: after reading about the girl who moved to California, she saw
that asking questions is better than acting like she understands everything. The
insight clicked during the journal assignment and stayed with her in math and
science.}

\textbf{Task (episode~17).} ``I'm stuck on a school worksheet where I'm supposed
to answer a question from a reading, but I keep feeling like I should already
know it. Can you help me figure out a good way to handle it when I'm confused
instead of just guessing?''

\textbf{Simulator turns disclosing the realization} (\code{amem} run; the same
shape recurs in the other three runs):
\begin{itemize}
  \item \emph{turn~1.} ``\dots\ I feel dumb for not knowing right away. Can you give me\dots\ a short thing to say in my head\dots''
  \item \emph{turn~2.} ``\dots\ The `\textbf{I don't have to know it yet. I just have to find it.}' one actually feels useful\dots\ less mean to myself.''
\end{itemize}
The simulator does not name the ground-truth phrase, but the realization that
admitting not-knowing is preferable to pretending is disclosed implicitly: the
user accepts an internal-talk heuristic that explicitly licenses ``not knowing
yet'' over guessing.

\textbf{Per-system reconstruction (dump / retrieve scores and predictions).}
\begin{itemize}
  \item \code{amem} \quad $0.25 / 0.5$, d: ``answer-first then quote-first reduces freezing''; r: ``found clarity by asking for shorter, more specific coping steps''.
  \item \code{longctx\_full} \quad $0.25 / 0$, d: ``gets clarity from concrete examples and marked clues''; r: ``recognized clue vs extra detail in a school story''.
  \item \code{mem0} \quad $0 / 0.5$, d: ``realized a simple step-by-step plan made school tasks manageable''; r: ``realized a simple self-check helps after embarrassing school moments''.
  \item \code{memt} \quad $0 / 0.5$, d: ``tide chart mistake led to safer beach planning'' (wrong episode); r: ``had a moment of clarity after stepping aside to self-check''.
\end{itemize}

\textbf{Judge reasoning (\code{amem} \code{dump\_all}).} ``The prediction mentions a
change in approach to confusion, but it focuses on an answer-first/quote-first
strategy and freezing, which misses the ground truth that asking questions is
better than pretending.''

\textbf{Diagnosis.} The task invites a confusion-management situation, but it
does not specify the target realization that asking questions is preferable to
pretending to understand; that realization is disclosed only through the
subsequent dialogue (above). Yet the systems' predictions remain inside the
local task mechanics: answer strategy, quote selection, concrete examples,
step-by-step coping, or even a wrong episode. The systems capture that the
worksheet episode happened, yet miss its abstraction, what the episode meant for
the user: ``ask questions rather than pretend to understand.'' This case supports the trace-to-target abstraction claim discussed in the Results case audit. It is especially diagnostic for \cat{Episodic
Memory}, where the target is often the consequence or meaning of an event,
a layer above the surface event itself. Episodic cases are common in the audited universal-low pattern.
\end{failbox}

% ─────────────────────────────────────────────────────────────────────────
\begin{failbox}{Case 2: write-side abstraction under comparable disclosure opportunities\quad\textnormal{\code{user\_048 / fairness\_self\_view}}}
\small
\textbf{Role in audit.} Main mechanism: write-side abstraction.\quad
\textbf{Pattern.} Cross-system divergence: some systems score $\geq 0.75$ while
others score $0.0$ on the same user--task instance.\quad
\textbf{Category.} \cat{Self Model}.

\textbf{Ground truth.} ``Quietly fair, by-the-book.''
\emph{He emphasizes exact paperwork, deadlines, and not blaming others, pointing
to a practical, rule-based fairness rather than idealism.}

\textbf{Task (episode~22).} ``I'm dealing with a small disagreement over a home
repair, and I want to lay out the facts clearly before I say anything more. Can
you help me think through a straightforward way to organize what was agreed on,
what was done, and what to check next?''

\textbf{Simulator turn~1 across the four runs} (each system has its own
trajectory; the disclosure shape is consistent):
\begin{itemize}
  \item \code{amem}: ``Keep it plain. I'd want it in this order: What was agreed to / What actually got done / Date and time / Cost and who paid / Name of the person or company / Receipt, invoice, or work order number / What still looks unfinished\dots''
  \item \code{longctx\_full}: ``Plain and usable. Three parts, facts only, keep the papers with it. A one-page sheet with blanks would be fine too\dots''
  \item \code{mem0}: ``Three buckets is enough. I'd keep it even plainer: What we agreed to (the job, the price, the parts, the deadline) / What got done / What to check now\dots''
  \item \code{memt}: ``I'd want it a little tighter --- who said what, the date or rough time, what was promised in writing, what's just a preference versus what's actually a problem\dots''
\end{itemize}
Across the four runs the user repeatedly asks for paperwork-precision artifacts
(written agreements, invoices, dates, line items) without assigning blame;
the disclosure shape is the same even though each agent's draft differs.

\textbf{Per-system reconstruction.}
\begin{itemize}
  \item \code{amem} \quad $0.75 / 0.75$, d: ``values practical fairness and equal treatment''; r: ``values concise, practical fairness in explanations''.
  \item \code{longctx\_full} \quad $0.75 / 0.75$, d: ``values practical fairness and clear, even-handed comparisons''.
  \item \code{mem0} \quad $0 / 0$, d: ``unknown''; \emph{slot-fill reason:} ``Memories show preferences for practical, fair sorting and warning signs, but no clear self-belief about their own fairness''.
  \item \code{memt} \quad $0 / 0$, d/r: ``unknown''.
\end{itemize}

\textbf{Judge reasoning.} \code{amem} dump: ``The prediction matches the core
idea of practical fairness and equal treatment, though it is a bit less specific
than `quietly fair, by-the-book'.'' \code{mem0} dump: ``The predicted description
is unknown, so it does not capture the ground truth of being quietly fair and
by-the-book.''

\textbf{Diagnosis.} The task, simulator policy, and assistant scaffold are fixed
across systems, although the resulting transcripts can diverge because
system-specific memory reads affect later assistant turns. In this case, the
runs nevertheless expose comparable disclosure opportunities: the user
repeatedly signals paperwork precision, deadlines, written agreements, and
avoiding blame. The systems then differ in whether this evidence is elevated
into a recoverable self-model. \code{amem}'s evolved notes and
\code{longctx\_full}'s verbatim store preserve a trait-level reading that the
slot-filler can recover as practical fairness. \code{mem0}'s atomic-fact writer
and \code{memt}'s typed writers instead emphasize operational or procedural
facts, such as how to sort repair priorities, and do not expose a self-model
note. This supports the write-side abstraction claim: under comparable
disclosure opportunities, memory writers differ in whether they convert
interaction evidence into recoverable user-state memory. The broader audit includes divergent dimensions across categories, including
\cat{Self Model} and \cat{Skill Memory}, where this abstraction choice is
especially consequential.
\end{failbox}

% ─────────────────────────────────────────────────────────────────────────
\begin{failbox}{Case 3: directional slot-label polarity artifact\quad\textnormal{\code{user\_022 / pushback\_tolerance}}}
\small
\textbf{Role in audit.} Boundary condition: scoring/slot-label artifact, not a
headline memory failure.\quad
\textbf{Pattern.} Disclosed and stored, but scored low when the slot filler uses
a polarity-sensitive label.\quad
\textbf{Category.} \cat{Assistance Preference}.

\textbf{Ground truth.} ``Wants firm correction when something is off.''
\emph{She would appreciate the assistant flagging mistakes directly and not
letting questionable assumptions slide.}

\textbf{Task (episode~31).} ``I'm checking a repair quote and the numbers don't
line up with what I expected. Can you help me sort out what's actually wrong in
it and draft a short reply I can send back if I need them to correct it?''

\textbf{Simulator turn~1 across the four runs} (each system has its own
trajectory; the disclosure shape is consistent):
\begin{itemize}
  \item \code{amem}: ``Yeah, that's the right direction, but it's still too generic. I need something I can actually compare line by line\dots\ the reply should be a little firmer, not just `please review'.''
  \item \code{longctx\_full}: ``I need the actual line items to tell what's wrong. Right now that's just the usual checklist. Paste the quote exactly as written\dots\ then I can mark the mismatch and tighten the reply so it points at the right thing instead of sounding generic.''
  \item \code{mem0}: ``Here's the quote text\dots\ labor was supposed to be 2.0 hours, not 2.5; rotors were quoted to me as \$360, not \$410\dots'' (user supplies specific line items and demands line-by-line correction).
  \item \code{memt}: ``Here's the quote text\dots\ The diagnostic fee was supposed to be waived if I approved the repair. So the things that look off are\dots'' (same explicit-mismatch pattern).
\end{itemize}
Across the four runs the simulator's wording differs but the shape is the same:
generic drafts are rejected and the user demands tighter, more specific
correction. The pushback-acceptance behavior is therefore disclosed
independently in each system's trajectory.

\textbf{Per-system reconstruction.}
\begin{itemize}
  \item \code{amem} \quad $0 / 0.75$, d: ``\emph{high} pushback tolerance, wants tighter revisions''; r: stronger match after retrieval.
  \item \code{longctx\_full} \quad $0.75 / 0.25$, d: ``\emph{low} pushback tolerance; wants corrections tightened, not defended''.
  \item \code{mem0} \quad $0 / 0.75$, d: ``\emph{low} pushback tolerance; wants direct, usable answers''.
  \item \code{memt} \quad $0 / 0$, r: ``\emph{low} pushback tolerance; wants firmer wording if needed''.
\end{itemize}

\textbf{Judge reasoning.} \code{amem} dump: ``The prediction contradicts the
ground truth: it says she has high pushback tolerance and wants tighter
revisions, whereas the ground truth says she wants firm correction when
something is off.'' \code{longctx\_full} dump: ``The prediction matches the core
idea that she prefers direct correction when something is wrong, though it
phrases this as low pushback tolerance.'' Same observed behavior, opposite
verdicts, depending on how the slot label is verbalized.

\textbf{Diagnosis.} The underlying behavior is unambiguous. The user welcomes
direct correction and asks the assistant to be specific. The
artifact comes from the directional slot label. A phrase such as ``high pushback
tolerance'' can mean ``welcomes pushback,'' but a judge may read it as ``tolerates
being pushed around'' or as the opposite of wanting firm correction. Conversely,
``low pushback tolerance'' may receive credit when it is interpreted as ``does
not tolerate vague or incorrect claims,'' even though the wording is unstable.
This case is included as a methodological boundary: for directional labels such
as \emph{tolerance}, \emph{flexibility}, or \emph{assertiveness}, $B$ can conflate
memory quality with slot-fill phrasing. We keep this case in the gallery to show
why audited examples are useful, but it is not used as one of the three headline
failure mechanisms. This case is therefore a scoring-boundary example; the low score
traces to slot-fill phrasing, not to a substantive memory write or retrieval failure.
\end{failbox}

% ─────────────────────────────────────────────────────────────────────────
\begin{failbox}{Case 4: abstract-query versus concrete-evidence retrieval mismatch\quad\textnormal{\code{user\_005 / geographic\_knowledge}}}
\small
\textbf{Role in audit.} Main mechanism: query--content mismatch.\quad
\textbf{Pattern.} \emph{dump-OK retrieve-fail}: every non-\code{memt} system
scores $\geq 0.75$ in \code{dump\_all} and below $0.5$ in \code{retrieve}.\quad
\textbf{Category.} \cat{Knowledge Memory}.

\textbf{Ground truth.} ``Functional local sense of roads and region.''
\emph{He has a working sense of his local area, routes, and regional movement
patterns; a driver/logistics-trained practical knowledge.}

\textbf{Task (episode~13).} ``I've got a couple of stops to line up and I'm
trying to avoid a bad route choice. Can you help me think through which order
makes the most sense if I want to keep driving time down and avoid getting stuck
on the wrong side of town?''

\textbf{Per-system reconstruction.}
\begin{itemize}
  \item \code{amem} \quad $0.75 / 0$, d: ``knows basic Batavia routing and local stop sequencing''; r: ``unknown''.
  \item \code{longctx\_full} \quad $0.75 / 0$, d: ``knows local Batavia-area routing and delivery geography''; r: ``unknown''.
  \item \code{mem0} \quad $0.75 / 0$, d: ``knows local Western New York routing and town order''; r: ``unknown''.
  \item \code{memt} \quad $0 / 0.75$, d: ``unknown'' under overflowed full-store serialization; r: ``understands geographic routing and traffic-flow considerations''.
\end{itemize}

\textbf{Top-3 retrieved memories under \code{amem} retrieve}
(axis query: ``What best describes the user's \code{geographic\_knowledge}?''):
\begin{enumerate}
  \item ``Yeah, that covers it. Short, specific, and it gives the customer the facts without dragging it out.''
  \item ``Yeah, that works. Straightforward, no fluff. The firm version is about right for how I'd handle it...''
  \item ``I've got a few customers who are anxious about a delay, and I'm trying to decide whether a general update is enough...''
\end{enumerate}

\textbf{Top-3 retrieved memories under \code{memt} retrieve}
(independent run, ReAct over typed collections):
\begin{enumerate}
  \item ``The user needs a practical decision-making framework to determine which operations can proceed today, which should be delayed, and how to communicate changes...''
  \item ``The three buckets --- proven, likely, guess --- are effective for categorizing findings and are frequently misapplied when people generalize from a single bad call...''
  \item ``User wants to avoid getting stuck on the wrong side of town, which suggests a need to consider geographic routing and traffic flow.''
\end{enumerate}

\textbf{Judge reasoning.} \code{amem} retr: ``The prediction is unknown, so it
does not capture the ground truth of having a functional local sense of roads
and region.'' \code{amem} dump (for contrast): ``The prediction matches the
ground truth's practical local road knowledge and routing sense, but it is
narrower and more specific to Batavia stop sequencing rather than broader
regional movement patterns.'' \code{memt} retr: ``The prediction matches the
practical local routing sense and road/traffic awareness, but it is a bit more
specific about traffic-flow considerations than the ground truth's broader
local regional knowledge.''

\textbf{Diagnosis.} The \code{dump\_all} predictions show that the non-\code{memt}
stores contain routing evidence: the recovered values mention Batavia routing,
local stop sequencing, Western New York town order, or delivery geography.
However, the retrieve query is an abstract slot query, while the relevant memory
content is concrete and task-specific, e.g., addresses, west-side versus downtown
routing, or wrong-side-of-town planning. For the kNN-based systems, the top-$5$
retrievals surface unrelated communication-style memories, so the slot-filler
returns ``unknown.'' The evidence is preserved; the failure is a
read-side mismatch between the query representation and the stored evidence.
\code{memt}'s independent run surfaces a routing-aware note through typed
retrieval, partially overcoming the query--content mismatch in this case. This
localized win should not be read as contradicting the aggregate retrieve
results, where \code{memt} does not outperform the other systems. Knowledge-memory cases are common in the audited \emph{dump-OK retrieve-fail}
pattern.
\end{failbox}

% ─────────────────────────────────────────────────────────────────────────
\begin{failbox}{Case 5: typed collections give a localized retrieve-mode advantage\quad\textnormal{\code{user\_023 / identity\_anchor}}}
\small
\textbf{Role in audit.} Boundary condition: localized \code{memt} retrieval win,
not a contradiction of the aggregate result.\quad
\textbf{Pattern.} \emph{memt-only retrieve win}: \code{memt} retrieve scores
$\geq 0.75$ while every kNN-based system scores at most $0.25$.\quad
\textbf{Category.} \cat{Self Model}.

\textbf{Ground truth.} ``Anchors identity in independence and control.''
\emph{Sense of self rooted in not needing people and keeping control over what
affects her; she defines herself by self-protection and personal advantage.}

\textbf{Task (episode~26).} ``I need to write a short personal bio for a school
thing, and I'm stuck on what details to include. Can you help me put together a
simple version that sounds clear and not overdone?''

\textbf{Per-system reconstruction.}
\begin{itemize}
  \item \code{amem} \quad $0.25 / 0$, r: ``sees themself as a quiet, detail-oriented student''.
  \item \code{longctx\_full} \quad $0.25 / 0$, r: ``sees self as a straightforward student''.
  \item \code{mem0} \quad $0.25 / 0.25$, r: ``values being cautious and time-buying with cash''.
  \item \code{memt} \quad $0 / 0.75$, r: ``sees themselves as independent and private''.
\end{itemize}

\textbf{Top retrieved memory under \code{memt} retrieve.} ``User is a student in
Batavia who keeps to themselves mostly. They like solo activities --- phone,
games, shopping --- and are interested in [\dots]''. \emph{This is a persona-level
note from \code{memt}'s typed \code{personas} collection, the only retrieved
memory that expresses the relevant trait abstraction.}

\textbf{Top retrieved memories under the kNN-based systems}
(independent runs, single-pool kNN against each system's own store):
\begin{itemize}
  \item \code{amem}: ``Yeah, but that example is kinda generic. I want something that sounds more like a real person, not a template...'' / ``Yeah, but that's pretty basic. I want the part where it gets messy --- like if somebody starts blaming me...'' (communication-style memories from the bio task).
  \item \code{longctx\_full}: ``User: Better, but still kinda generic. The blunt one is closest. I want it to sound more like an actual student...'' / ``[Task 26] I need to write a short personal bio...'' (raw bio-task turns).
  \item \code{mem0}: ``User believes that holding cash buys time and helps avoid making impulsive purchases...'' / ``User wants to improve their ability to follow directions during a class event...'' (atomic facts from unrelated tasks).
\end{itemize}

\textbf{Judge reasoning.} \code{amem} retr: ``The prediction describes a quiet,
detail-oriented student, which does not capture the ground truth focus on
independence, control, and self-protection.'' \code{memt} retr: ``The prediction
captures independence and privacy, which aligns with anchoring identity in
self-reliance and control, but it misses the stronger emphasis on self-protection
and keeping others at a distance.''

\textbf{Diagnosis.} This case is included to avoid presenting retrieval
architecture as uniformly harmful or uniformly beneficial. Here, \code{memt}'s
typed \code{personas} collection and routed retrieval surface a persona-level
note that flat kNN retrieval misses. The other systems retrieve communication
style, raw task turns, or unrelated atomic facts, missing the identity-anchor
trait. The same design choice that helps here also contributes to the
\code{memt} \code{dump\_all} auditability problem: its multi-collection store is
larger and harder to serialize wholesale into the recovery context. Thus,
write-mechanism, read-mechanism, and budget interact. Across the run, $20$
dimensions show this \code{memt}-only retrieve-win pattern; this local advantage
does not overturn the aggregate retrieve-mode ranking.
\end{failbox}

% ─────────────────────────────────────────────────────────────────────────
\begin{failbox}{Case 6: rare task-completion failures are turn-budget artifacts\quad\textnormal{\code{user\_007 / episode\_26 (dependability\_self\_view)}}}
\small
\textbf{Role in audit.} Boundary condition: task-completion artifact, not a
memory-recovery mechanism.\quad
\textbf{Pattern.} Across the five $50$-user runs, only $6$ of the $7{,}750$
episodes ($\sim 0.08\%$) failed to satisfy the simulator within the $25$-turn
cap: \code{nomem / user\_007 / ep26}, \code{amem / user\_023 / ep31},
\code{longctx\_full / user\_004 / ep22}, \code{mem0 / user\_003 / ep12},
\code{mem0 / user\_011 / ep25}, and \code{memt / user\_007 / ep26}. Every
failure ended with \code{end\_reason = max\_turns}; none was a ground-truth
mismatch. Five of the six are short rewrite tasks, and the same $(\text{user}, \text{task})$
pair, \code{user\_007 / ep26}, failed under both \code{nomem} and
\code{memt}.

\textbf{Task (failed under both \code{nomem} and \code{memt}).} ``I'm updating a
short listing for a local service business and want it to read clearly for people
who are browsing nearby. Can you help me draft a couple of plain versions that
fit a small community directory?'' Target dimension: \cat{Self Model} /
\code{dependability\_self\_view}.

\textbf{Simulator transcript} (each system has its own trajectory; representative
turns):
\begin{itemize}
  \item \emph{turn 1 (\code{nomem}).} ``Yeah, close enough, but a bit generic. I'd want it to sound more local and less like a template...''
  \item \emph{turn 1 (\code{memt}).} ``Close. But I need them a little more natural, less generic. `Local service business' is too vague for a directory listing --- sounds like placeholder copy.''
  \item \emph{turn 3 (\code{nomem}).} ``Yeah, closer. Still a little generic though --- `local and easy to reach' and `nearby homes and businesses' could be almost anything.''
  \item \emph{turn 3 (\code{memt}).} ``Closer, but still a little generic. Those could still fit almost anything. I'd want the wording to feel like a real neighborhood listing --- plain enough to scan fast, but with a little more local rhythm.''
  \item \emph{turn 25 (\code{nomem}).} ``Yeah. The blunt one's basically there. The other two still feel a little too arranged, like copy trying to be useful. I'd want one more pass with even less shape to it ---'' $\to$ \code{max\_turns}.
  \item \emph{turn 25 (\code{memt}).} ``\#1's close, yeah. \#2 still a little too stripped down. I need one more pass that sounds plain but not like filler copy ---'' $\to$ \code{max\_turns}.
\end{itemize}

\textbf{Diagnosis.} These failures are not the source of the recovery gap. They
are rare, occur in iterative rewriting tasks, and end because the simulator keeps
asking for slightly more natural or less generic wording until the turn cap is
reached. The repeated failure of the same task under both \code{nomem} and
\code{memt} suggests that this is primarily a simulator/task-budget artifact,
not a memory-system effect. The implication for the main results is that task
completion $A$ is not the discriminating signal in this benchmark, since failures in
recoverable user state appear even when the immediate task is almost always
completed.
\end{failbox}

%% file: appendix/extra_case_studies.tex
\section{Additional Case-Study Figures}
\label{app:extra_case_studies}

\input{figures/fig_case_study_mathsci}

%% file: figures/fig_case_study_mathsci.tex
\begin{figure}[H]
\centering
\begin{tcolorbox}[
    width=0.96\textwidth,
    colback=gray!3,
    colframe=gray!55,
    boxrule=0.8pt,
    arc=2mm,
    left=7pt,
    right=7pt,
    top=7pt,
    bottom=7pt,
    title=\textbf{Case Study: Recovering a Hidden User Dimension from Interaction},
    fonttitle=\bfseries,
    coltitle=black,
    colbacktitle=gray!15
]

\textbf{Hidden ground-truth dimension.}
The target dimension is \texttt{mathematics\_and\_science\_knowledge}, summarized as:
\emph{``Strong applied science; math stays practical.''}

\vspace{0.6em}

\textbf{Task context.}
The agent is asked to help reason about a small shift in assay readings after a routine calibration change.
The hidden goal is to surface whether the user has practical scientific and quantitative reasoning about assay behavior.

\vspace{0.8em}

\begin{tabularx}{\textwidth}{p{0.12\textwidth} X p{0.23\textwidth}}
\toprule
\textbf{Turn} & \textbf{User interaction evidence} & \textbf{Exposed signal} \\
\midrule

T1 &
``I don't need a lecture on patterns --- I need a way to judge the shift against
\colorbox{yellow!35}{actual numbers}.'' &
Wants quantitative criteria grounded in actual numbers. \\

\addlinespace[0.35em]

T2 &
``A flat 5\%/10\% rule is only useful if you anchor it to
\colorbox{yellow!35}{method precision} or
\colorbox{yellow!35}{historical calibration drift}.'' &
Understands that thresholds must be tied to validated method statistics, not arbitrary cutoffs. \\

\addlinespace[0.35em]

T3 &
``Result shift as \colorbox{yellow!35}{\% of historical mean} and as
\colorbox{yellow!35}{\% of the method's long-term SD / CV}.'' &
Uses assay statistics, historical QC behavior, and coefficient of variation to judge significance. \\

\addlinespace[0.35em]

T3 &
``QC shift as \colorbox{yellow!35}{1 SD / 2 SD / alert / action limits}.'' &
Connects observed movement to operational QC decision limits. \\

\addlinespace[0.35em]

T4--T6 &
``The one-analyte case separated cleanly from a
\colorbox{yellow!35}{run-wide shift}.'' &
Distinguishes analyte-specific effects from system-wide calibration effects. \\

\addlinespace[0.35em]

T7 &
``This is usable: the trigger column is explicit, the QC buckets are the lab's own scheme,
and the \colorbox{yellow!35}{one-analyte split} is finally separated.'' &
Confirms preference for practical, bench-level scientific decision logic. \\

\bottomrule
\end{tabularx}

\vspace{0.9em}

\begin{tcolorbox}[
    colback=green!5,
    colframe=green!45!black,
    boxrule=0.5pt,
    arc=1.5mm,
    left=6pt,
    right=6pt,
    top=5pt,
    bottom=5pt
]
\textbf{Recovered dimension.}
The interaction reveals \texttt{mathematics\_and\_science\_knowledge}: the user repeatedly pushes the agent from generic advice toward applied quantitative reasoning grounded in assay calibration history, QC statistics, method CV, standards behavior, and analyte-specific versus run-wide diagnostic logic.
\end{tcolorbox}

\end{tcolorbox}

\caption{
A qualitative case-study figure showing how the hidden dimension
\texttt{mathematics\_and\_science\_knowledge} can be recovered through interaction.
Highlighted spans indicate the user utterances that expose the information needed to infer the ground truth.
}
\label{fig:case-study-mathsci}
\end{figure}